\documentclass[lettersize,journal]{IEEEtran}
\usepackage{times}
\usepackage{multirow}
\usepackage{algorithmic}
\usepackage{algorithm}
\usepackage{array}
\usepackage{amsmath}
\usepackage{amssymb}
\usepackage{graphicx}
\usepackage[table]{xcolor}
\usepackage{colortbl}
\newcommand{\mycc}[1]{\cellcolor{yellow!30}#1}
\usepackage{stfloats}
\usepackage{url}
\usepackage{verbatim}

\usepackage{cite}
\usepackage{float}
\usepackage{multicol}
\usepackage{hyperref}
\hypersetup{
    colorlinks=true,
    linkcolor=blue,
    filecolor=magenta,      
    urlcolor=cyan,
    pdftitle={Overleaf Example},
    pdfpagemode=FullScreen,
    }

\begin{document}

\title{Can Foundation Models Generalise the Presentation Attack Detection Capabilities on ID Cards?}

\author{
    Juan~E.~Tapia,~\IEEEmembership{Senior Member,~IEEE,}
    and~Christoph~Busch,~\IEEEmembership{Senior~Member,~IEEE}%
\thanks{Juan Tapia and Christoph Busch are with the da/sec-Biometrics and Internet Security Research Group, Hochschule Darmstadt, Germany, e-mail: \{\href{juan.tapia-farias@h-da.de}{juan.tapia-farias}, \href{christoph.busch@h-da.de}{christoph.busch}\}@h-da.de.}%
\thanks{Manuscript received Month DD, YYYY; revised Month DD, YYYY.}}

\markboth{Journal of \LaTeX\ Class Files,~Vol.~14, No.~8, August~2021}%
{Shell \MakeLowercase{\textit{et al.}}: A Sample Article Using IEEEtran.cls for IEEE Journals}


\maketitle

\begin{abstract}
Nowadays, one of the main challenges in presentation attack detection (PAD) on ID cards is obtaining generalisation capabilities for a diversity of countries that are issuing ID cards. Most PAD systems are trained on one, two, or three ID documents because of privacy protection concerns. As a result, they do not obtain competitive results for commercial purposes when tested in an unknown new ID card country. In this scenario, Foundation Models (FM) trained on huge datasets can help to improve generalisation capabilities. This work intends to improve and benchmark the capabilities of FM and how to use them to adapt the generalisation on PAD of ID Documents. Different test protocols were used, considering zero-shot and fine-tuning and two different ID card datasets. One private dataset based on Chilean IDs and one open-set based on three ID countries: Finland, Spain, and Slovakia. Our findings indicate that bona fide images are the key to generalisation.
\end{abstract}


\section{Introduction}
\IEEEPARstart{A} traditional biometric remote verification system comprises two components: one is related to identifying whether the data subject in front of the capture device is ``live," which means physically present in front of the smartphone; images captured in this scenario are not controlled by an operator and therefore represent ``wild" conditions. The second component detects if the ID Card presented by the subject is the original one issued by the official governmental service, which means without any adulteration or modified information. 

In order to evaluate the authenticity of the document, a Presentation Attack Detection (PAD) component is used in the system. Typically, most of the time, the PAD system detects bona fide images, which are the pristine images that a subject presents in front of the capture device and different types of attacks. 

The most common attacks are:  The "printed" attacks which are ID Cards printed on glossy paper. The ``Screen" attacks are images displayed on a secondary smartphone, TV, or laptop screen. Further, the ``Border" attack represents a modified ID Card created by copy-paste and swapping different areas from one ID Card to another to impersonate an identity and obtain benefits. 

Nowadays, solving this challenge is very difficult because it is very asymmetrical. The attacker only needs one image, showing a very well-created fake ID document, to fool the system. This attack can be executed based on digital tools available today, instead of thousands of ID cards with bona fide images that the defenders/companies need to train a robust PAD classifier based on deep learning or vision transformer models. 

This challenge is even harder based on the privacy protection of ID cards and passports concerns. Because of that, the defender models have a complex challenge to solve and satisfy the need for generalisation capabilities. There is no available universal PAD system that can be trained based on a dataset from a few countries and detect different attacks from any other country with a low error rate. 

Motivated by this challenge, many researchers have studied and developed methods to adapt the existing PAD on ID cards to another domain or to another country while re-training with a reduced number of samples of the new country supported by meta-learning techniques, such as one-shot, few-shot, or zero-shot approaches \cite{survey}.

Furthermore, there are apparent differences in ID documents from different countries, as not all ID Cards or ID documents are ICAO compliant. This lack of standardisation presents a fundamental problem to solve. 

Documents from different countries present different challenges: from Asian countries such as China, Japan, South Korea, Thailand, Malaysia and others, the ID card presents symbols instead of regular Latin or Greek letters or a combination of both. The same problem is presented for the Arabic alphabet. This factor makes the applicability of the system across the world challenging to obtain. 

The material exhibits significant variability when it comes to PVC/plastic ID cards and passports. For example, the printed Visa version used in the USA differs from a plastic document used in Chile, and plastic laminated like the Peruvian ID card. This variability can create confusion for PAD systems. For instance, plastic laminated documents are often flagged as potential threats.

Motivated by the previous challenges, we propose an adaptation method and a benchmark performance analysis of the Foundation model as a downstream model trained with millions of images to be used in PAD on ID Card systems. \\

The main contributions of this work are:

\begin{itemize}
    \item A detailed benchmark is developed in order to study the generalisation capabilities of FM on PAD on the ID cards system on two datasets.
    \item An efficient method based on a fine-tuning approach from FM is proposed, which improves the results with a zero-shot approach.
    \item A second method based on concatenation of FM and a deep learning approach at the score level is proposed, which can boos the results in the open-set ID-Net dataset.
    \item We demonstrated that the key for generalisation capabilities is to have representative bona fide images instead of thousands of attacks and looking for local and global features on ID cards.
\end{itemize}

The rest of the paper is organised as follows: Section \ref{sec:related} reviews the related works. Section \ref{sec:data} shows and describes all the datasets used in this work. Section \ref{sec:metrics} explains the metrics used to analyse the experiments. Section \ref{sec:results} explains the experiments and results obtained, and Section \ref{sec:conclusion} draws the findings of this work.

\section{Related work}
\label{sec:related}

\subsection{PAD on ID Cards}
Nowadays, the PAD on ID Cards is a relevant topic based on the increasing attempts to fool remote biometric systems. Remote system applications have been spread in several areas, such as banking, onboarding, migrations, certification requests, and many others. 

The PAD on ID cards is a challenging topic to move forward because there is a limited quantity of datasets available to train deep learning or vision transformer classifiers, which represent the state of the art today. The performance is related to the number of images available, which is limited by data protection regulations.

Today, we have open-set datasets available that present some limitations to training a robust PAD. These datasets are not realistic because they are not ICAO-compliant, thus they do not represent the real case \cite{tapia2025pass}. For instance, many face images on ID cards present pose rotation, hat, sunglasses or poor resolution, which is not a real operational case \cite{Merkle-OFIQ-Report-240930, Yang-FaceImageQuality-SIDD-NISTIR8485-2024}. 

The best results on state-of-the-art data have been demonstrated by private ID Card datasets, which show a handicap in performance based on genuine bona fide images since these datasets are not available due to privacy restrictions \cite{GONZALEZ-PR}.

In the literature, a few open-access datasets are available, such as MIDV500 \cite{MIDV500}, MIDV2020 \cite{MIDV2020}, DLC2021 \cite{MIDVDLC2021}, MIDV-Holo \cite{MIDVHolo}, and IDNet \cite{IDNet}. Most of them are created from a few subjects with many images, different attacks, and a small number of countries. 

A remarkable effort was proposed by the IDNet dataset, which comprises 837,060 images of synthetically generated identity documents, totalling approximately 490 gigabytes, categorised into 20 types from 10 U.S. states' driver's licenses and 10 European passports and ID cards. However, stronger and more reliable results have been reported by private datasets with a high number of bona fide subjects, documents, attacks and images per subject.

Gonzalez et al.\cite{Gonzalez-Tbiom} proposed one of the two-stage presentation attack detection methods for Chilean ID cards using a private dataset based on MobileNetV2. The same authors extended this proposal using different capture sources, and several presentation attack instruments such as printing, display, composite, plastic (PVC), and synthetic ID card images \cite{GONZALEZ-PR} using a database consisting of 190,000 real-case Chilean ID card images, which was also supported by a private company. 

Benalcazar et al. \cite{benalcazar} proposed one of the first methods to generate synthetic ID cards based on a Generative Adversarial Network (GANs), showing the limitations of this technique to create new images. This work also used a Chilean ID card from a private dataset.

Markam et al. \cite{Markham} proposed a method to create attacks of printed and screen images using a transfer style algorithm exploring CycleGANs \cite{CycleGAN2017} and Pix2pixHD \cite{isola2017image}. The technique was evaluated considering only open-set datasets such as MIDV-2020 and DLC-2021 datasets.

Rocamora et al. \cite{Alvaro} proposed a Few-Shot method based on a prototypical network in order to improve the generalisation capabilities, including a small set of new ID cards from some countries that are not included in the training set. This protocol and method simulates how to extend the detection capabilities to a new ID card country with a reduced set of samples.

Tapia et al. \cite{Tapia-IJCB2024} organised the first competition on PAD on ID cards held at the International Joint Conference on Biometrics (IJCB) in 2024, which shows and remarks the generalisation capability challenge. The winner team method still shows a critical gap when evaluated with an unknown dataset of four different countries. This competition considers bona fide, composite, print and screen attacks. Most of the participating teams reached a good generalisation to only one or two countries. These results highlight the challenge of this topic and show that they are not one universal PAD that can detect any ID card. The quantity of training data is very relevant.

Chen et al. \cite{Chen-tifs} have developed a method focusing on one specific attack, such as printed ID documents. They proposed a document recapture detection scheme by employing a Siamese network to compare and extract distinct features in a recaptured document image. The same author proposes in \cite{chen-spectral} a method based on Moire pattern detection in the digital domain.

\begin{table*}[]
\centering
\scriptsize
\caption{Nº Img represents the total number of documents. Nº users (number of different documents). The template represents a description of the nature of the template or bona fide. Types of attacks: PVC represents and ID card printed on a plastic. ICAO represents whether the dataset follows the standard.
IDC: ID card, DLI: Driver's license, PSP: Passport, FOC: Resident permit}
\label{tab:survey_Idcard}
\begin{tabular}{|c|c|c|c|c|c|c|}
\hline
Database &
  \begin{tabular}[c]{@{}c@{}}Nª\\ Images\end{tabular} &
  \begin{tabular}[c]{@{}c@{}}Nº\\ Users\end{tabular} &
  \begin{tabular}[c]{@{}c@{}}Type of \\ Attacks\end{tabular} &
  \begin{tabular}[c]{@{}c@{}}Type of\\ Document\end{tabular} &
  \begin{tabular}[c]{@{}c@{}}Generate or\\ Capture \\ from template\end{tabular} &
  \begin{tabular}[c]{@{}c@{}}ICAO\\ Compliant\end{tabular} \\ \hline
\begin{tabular}[c]{@{}c@{}}KID34K \\ \cite{KID34K}\end{tabular} &
  $\sim$35K &
  82 &
  \begin{tabular}[c]{@{}c@{}}PVC, \\ Screen,\end{tabular} &
  \begin{tabular}[c]{@{}c@{}}IDC,\\ DLI\end{tabular} &
  Yes &
  No \\ \hline
\begin{tabular}[c]{@{}c@{}}MIDV\_500\\ \cite{MIDV500}\end{tabular} &
  $\sim$10K &
  50 &
  PVC &
  \begin{tabular}[c]{@{}c@{}}IDC,\\ DLI, \\ PSP, \\ FOC\end{tabular} &
  Yes &
  Partially \\ \hline
\begin{tabular}[c]{@{}c@{}}MIDV2019 \\ \cite{MIDV2019}\end{tabular} &
  $\sim$6K &
  50 &
  PVC &
  \begin{tabular}[c]{@{}c@{}}IDC,\\ DLI,\\ PSP,\\ FOC\end{tabular} &
  Yes &
  Partially \\ \hline
\begin{tabular}[c]{@{}c@{}}MIDV2020\\ \cite{MIDV2020}\end{tabular} &
  $\sim$72K &
  1k &
  PVC &
  \begin{tabular}[c]{@{}c@{}}IDC,\\ PSP\end{tabular} &
  Yes &
  Partially \\ \hline
\begin{tabular}[c]{@{}c@{}}MIDVDLC-2021 \\ \cite{MIDVDLC2021}\end{tabular} &
  $\sim$79K &
  80 &
  \begin{tabular}[c]{@{}c@{}}PVC, \\ Screen, \\ Print,\end{tabular} &
  \begin{tabular}[c]{@{}c@{}}IDC,\\ PSP\end{tabular} &
  Yes &
  Partially \\ \hline
MIDV-Holo \cite{MIDVHolo} &
  $\sim$30K &
  100 &
  \begin{tabular}[c]{@{}c@{}}PVC, \\ Digital\\ tampering\end{tabular} &
  \begin{tabular}[c]{@{}c@{}}IDC,\\ PSP\end{tabular} &
  Yes &
  Partially \\ \hline
\begin{tabular}[c]{@{}c@{}}Synt-ID-Card \\ \cite{benalcazar}\end{tabular} &
  $\sim$4K &
  - &
  Synthetic &
  IDC &
  No &
  Yes \\ \hline
IDNet \cite{IDNet} &
  $\sim$600K &
  $\sim$120K &
  \begin{tabular}[c]{@{}c@{}}Digital\\ tampering\end{tabular} &
  \begin{tabular}[c]{@{}c@{}}IDC,\\ PSP,\\ DLI\end{tabular} &
  Yes &
  Partially \\ \hline
SIDTD \cite{SIDTD} &
  $\sim$8K &
  1K &
  \begin{tabular}[c]{@{}c@{}}Digital\\ tampering,\end{tabular} &
  \begin{tabular}[c]{@{}c@{}}IDC,\\ PSP\end{tabular} &
  Yes &
  Partially \\ \hline
FMIDV \cite{FMIDV} &
  $\sim$28K &
  1K &
  \begin{tabular}[c]{@{}c@{}}Digital \\ tampering\end{tabular} &
  \begin{tabular}[c]{@{}c@{}}IDC,\\ PSP\end{tabular} &
  Yes &
  Partially \\ \hline
\begin{tabular}[c]{@{}c@{}}Gonzalez et al.\\ \cite{Gonzalez-Tbiom}\end{tabular} &
  10K &
  $\sim$30K &
  \begin{tabular}[c]{@{}c@{}}Border,\\ Print,\\ Screen\\ PVC\end{tabular} &
  IDC &
  No &
  Yes \\ \hline
\begin{tabular}[c]{@{}c@{}}Gonzalez et al.\\ \cite{GONZALEZ-PR}\end{tabular} &
  70K &
  $\sim$220K &
  \begin{tabular}[c]{@{}c@{}}Border,\\ Print,\\ Screen\\ PVC\end{tabular} &
  IDC &
  No &
  Yes \\ \hline
\end{tabular}%
\end{table*}

\subsection{Foundation models}
A strong interest has been observed in applying a foundation model for biometrics task \cite{fmbiometrics2025survey}, specifically for face recognition \cite{chettaoui}, presentation attack detection on face \cite{Ozgur_2025_WACV} and iris images \cite{tapia2025irisfm}.

The most adapted models among the state-of-the-art of foundation models are DinoV2 and CLIP. Both models were trained on two different huge datasets using self-supervised learning. 

DinoV2 \cite{dinov2}was trained on the LVD-142M dataset by retrieving images that are close to those in several curated datasets from a large pool of uncurated data. This VisionTransformer model has been focused on computer vision tasks, specifically on segmentation \cite{iris-sam}, but can be adapted to other visual functions with great success. 

Similarly, the Contrastive Language-Image Pre-training (CLIP) \cite{cherti2023reproducible, Radford2021LearningTV} was trained on a vast dataset of 400 million images. The CLIP model considers a branch for text and another branch for a visual task. 

Several applications have been proposed based on these foundation models. 
Shahreza et al. \cite{fmbiometrics2025survey} elaborate a complete survey on applying foundation models and large language models on biometrics and presentation attack detection.

Chettaoui et al. \cite{chettaoui} analyse whether the foundation models are ready to be applied to a face recognition system using Dinov2 and CLIP, showing the capabilities to improve in this context. 

Ozgur et al. \cite{Ozgur_2025_WACV} proposed the "FoundPAD" method, which adapts a CLIP model using LoRa Rank \cite{Lora} to improve face presentation attack detection. The author evaluates the generalisation benchmarks on diverse datasets, training pipelines for limited and synthetic data scenarios.

\section{Method}

For our proposed method, three concepts are relevant in order to understand our approach: Foundation models, Fine-tuning and Zero-shot, which are explained as follows:

Researchers introduced the term "foundation model" to describe machine learning models that are trained on a wide range of generalised and unlabeled data \cite{fmbiometrics2025survey}. These models are capable of performing various general tasks, because they were trained on vast datasets that include understanding language, generating text and images, and engaging in natural language conversations. On this paper, we are focusing on the visual branch of FM instead of text prediction capabilities.

The term zero-shot for image classification\footnote{\url{https://huggingface.co/docs/transformers/main/en/tasks/zero_shot_image_classification}} is a task that involves classifying images into different categories using a pre-trained network that has not been explicitly trained on labelled examples from those specific classes.  The deep learning and vision transformer model typically requires fine-tuning to recalibrate its knowledge.
These models learn aligned vision-language representations, which can be applied to a variety of downstream tasks, including zero-shot image classification on PAD.

The term "fine-tuning" in deep learning is a type of transfer learning technique. It involves taking a pre-trained network, which has already been trained on a huge dataset (such as DinoV2 and CLIP) for a general task like image recognition or natural language understanding, and making minor weight adjustments, unfreezing some layers to modify its internal parameters or adding an external classifier. The goal is to enhance the model’s performance on a new, related task without having to start the training process from scratch \cite{fmbiometrics2025survey, Chen-tifs}. 



In this work, we used a neural network as a head to fine-tune CLIP and DinoV2 and deep learning models, which allows it to efficiently produce a final embedding space adapted to the downstream PAD task, which does not depend on the internal adjustments of the models. In this scenario, the FM works as a feature extractor on top of which classification can be performed based on the image division performed by the vision transformer model inside the CLIP and DinoV2 architecture. Thus, we can explore features at different patch sizes. An additional fully connected layer with one output neuron, followed by a sigmoid layer, is added to the FM, resulting in the complete detector architecture. These layers are trained along with fine-tuning, using the binary cross-entropy loss $\mathcal{L}$, where ${\Upsilon}$ and $\hat{\Upsilon}$ represent the sample’s ground truth and
predicted labels, respectively.

\begin{equation}
    \mathcal{L} = -(\Upsilon_i\cdot \mathrm{log} \hat{\Upsilon}_i + (1 - \Upsilon_i)\cdot\mathrm{log}(1 - \hat{\Upsilon}_i))
\end{equation}

\begin{figure*}[]
\centering
\includegraphics[scale=0.45]{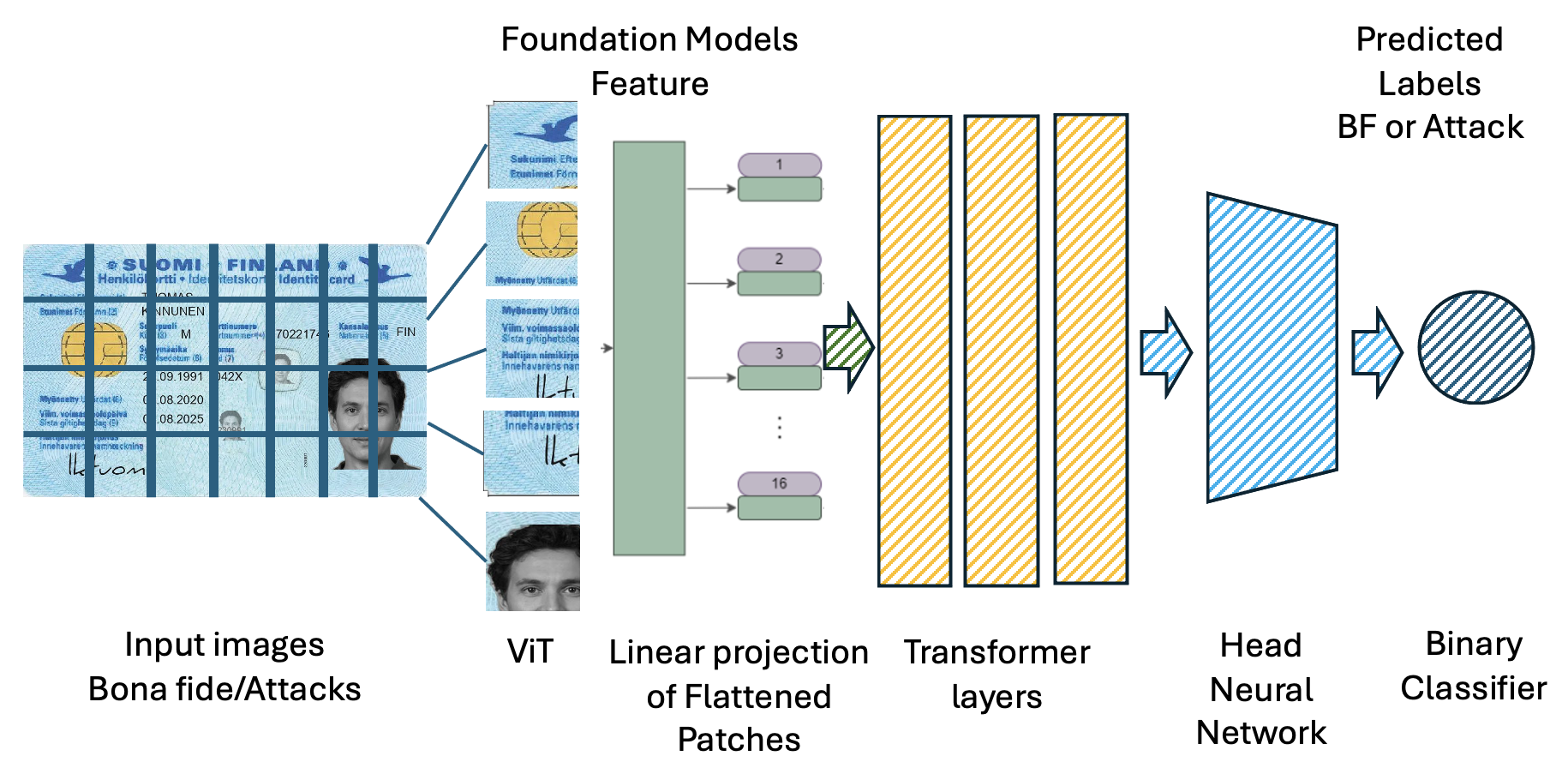}%
\caption{ID card Presentation Attack Detection pipeline. The ViT network divides the image into patches. The embedding space of the FM is adapted by fine-tuning using a head of the neural network to adapt parameters, and the classification layers are simultaneously trained to produce the PAD predictions. In zero-shot learning, the head neural network is not used.}
\label{fig:modelo}
\end{figure*}

Figure \ref{fig:modelo} illustrates an example of the PAD on the ID Card pipeline.
Furthermore, in order to take advantage of these foundation models and complement the traditional features from traditional deep learning, we additionally propose a second pipeline to concatenate at the score level the best results to improve the generalisation. The new pipeline is described in section \ref{improve}.

\section{Dataset}
\label{sec:data}
Two datasets that represent most of the cases available in the state of the art were explored to analyse the generalisation capabilities. 

First one, a private Chilean ID card dataset (CHL-DBP) \cite{Gonzalez-Tbiom} with four classes is used: bona fide, border, printed, and screen. On this dataset, the bona fide represents a genuine image present in front of the capture device by the owner. A border attack is an image that represents a change of information, copied and pasted from one ID card to another. This operation left traces and borders on the edges of the papers. A printed attack represents the bona fide image printed and recaptured using a smartphone and a monitor. The screen attack represents a replay attack for the ID card present on the screen of a mobile device, such as a tablet, laptop screen, or similar screen device.

The second one is an open-set dataset, ID-Net \cite{IDNet}, based only on an ID card subset containing ID Card images from Finland, Spain, and Slovakia. On this dataset, the "bona fide" images were generated from an empty template and are complete with synthetic information based on Stable Diffusion models. For the attacks, a "combined ID card" attack instrument species was included, which is an attack that represents changes and crop in the text and/or in the photo face area. The "Face-swap" represents attacks where the face photo is swapped between two ID cards. The "copy-move" eventually crops a random field from one identity document and moves it to another document. The "Face morphing" represents an attack where the image used to generate the fake ID card was morphed with another identity\footnote{Most details about the attacks of ID-Net may be found directly in the publication \cite{IDNet}}. Both datasets will be available only by request for research purposes only.

\begin{figure*}[]
\centering
\includegraphics[scale=0.09]{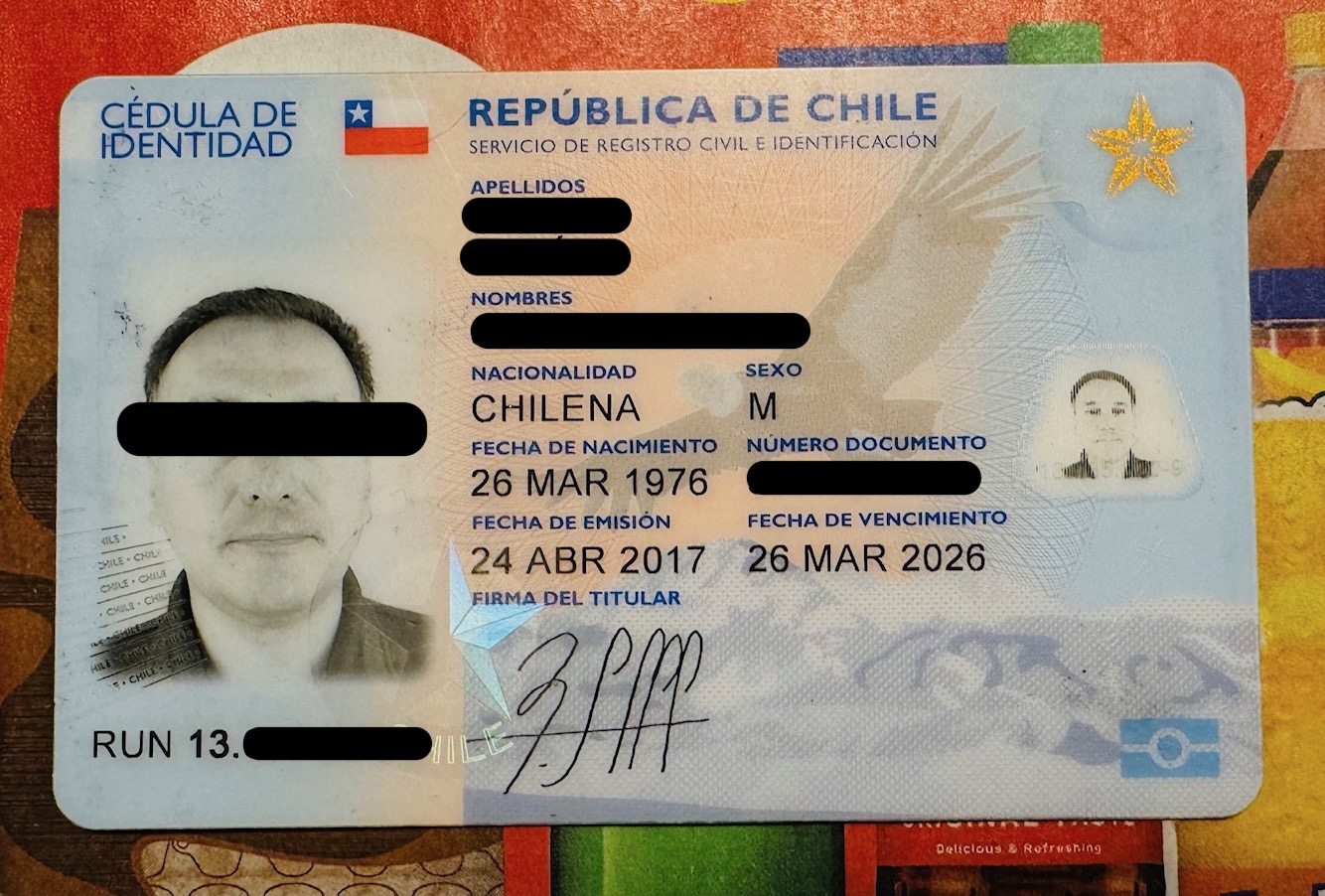} 
\includegraphics[scale=0.083]{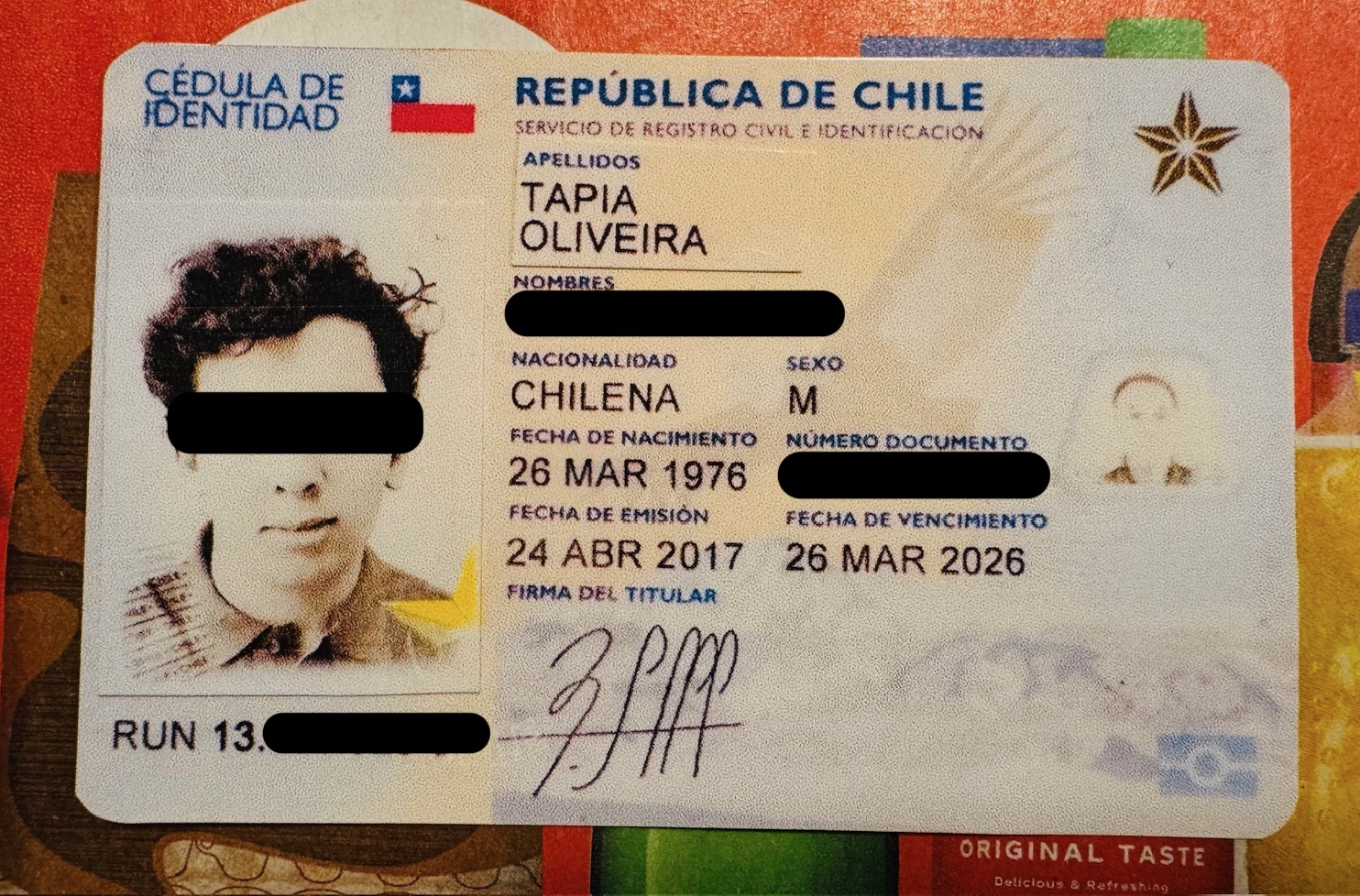} 
\includegraphics[scale=0.09]{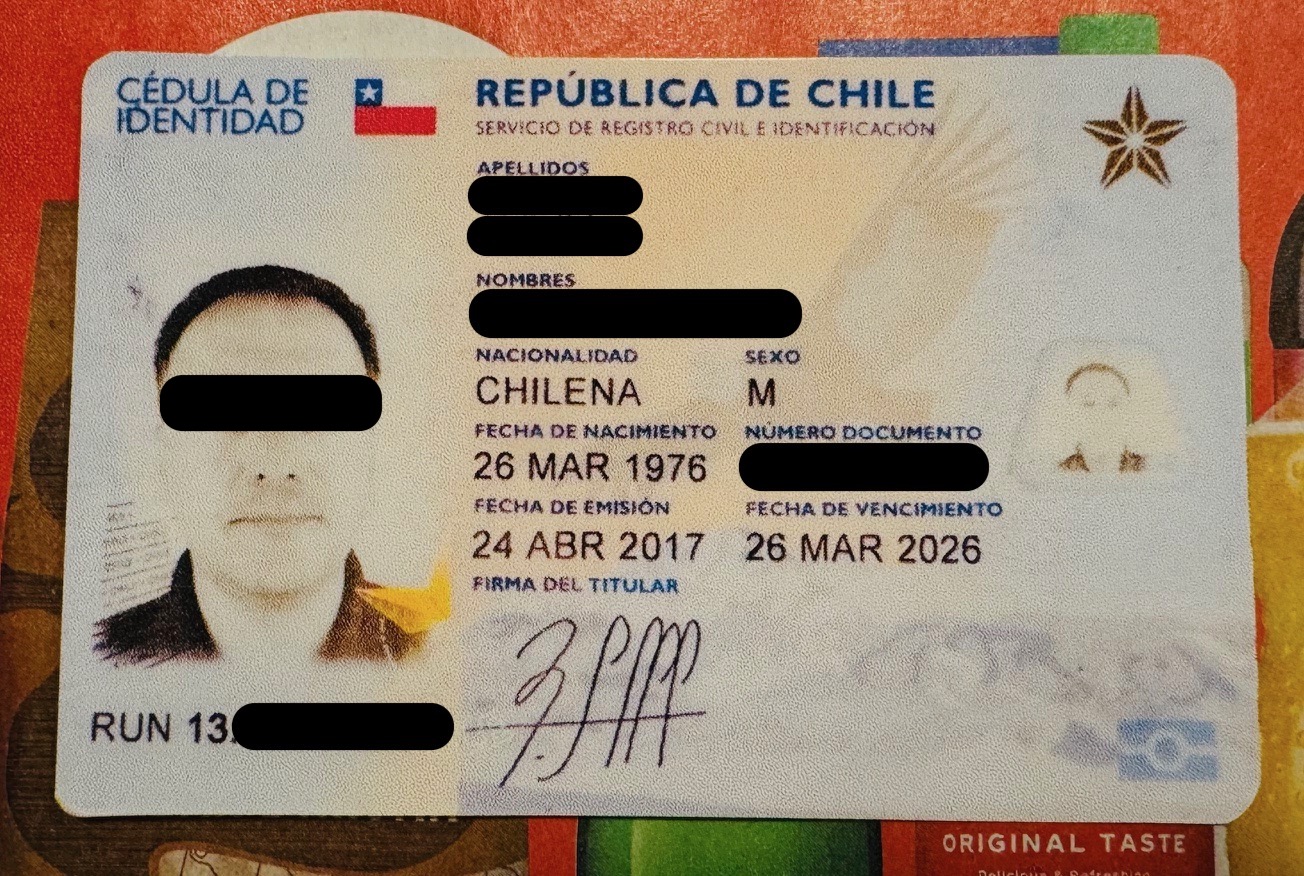}
\includegraphics[scale=0.08]{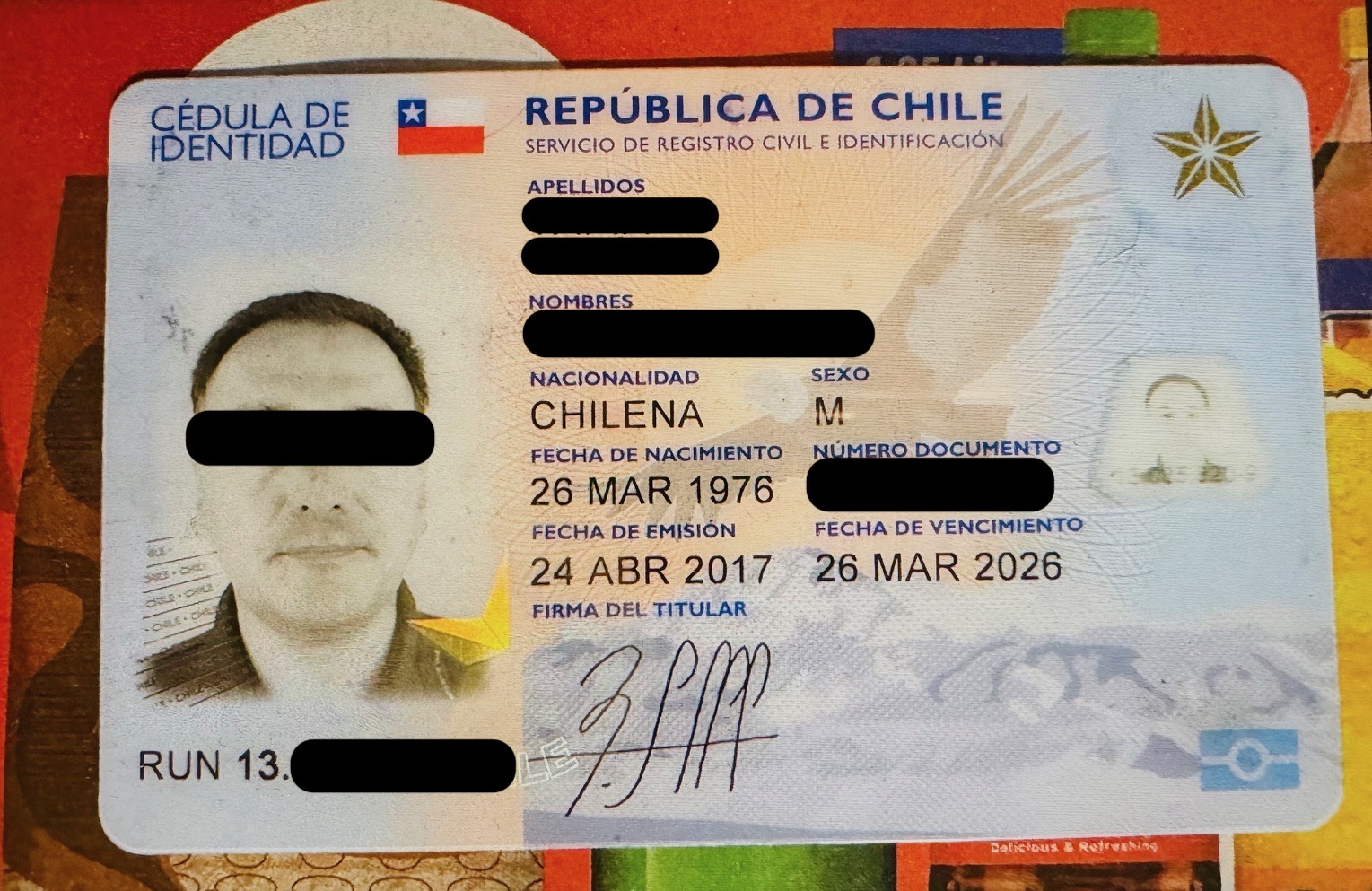} \\
\includegraphics[scale=0.13]{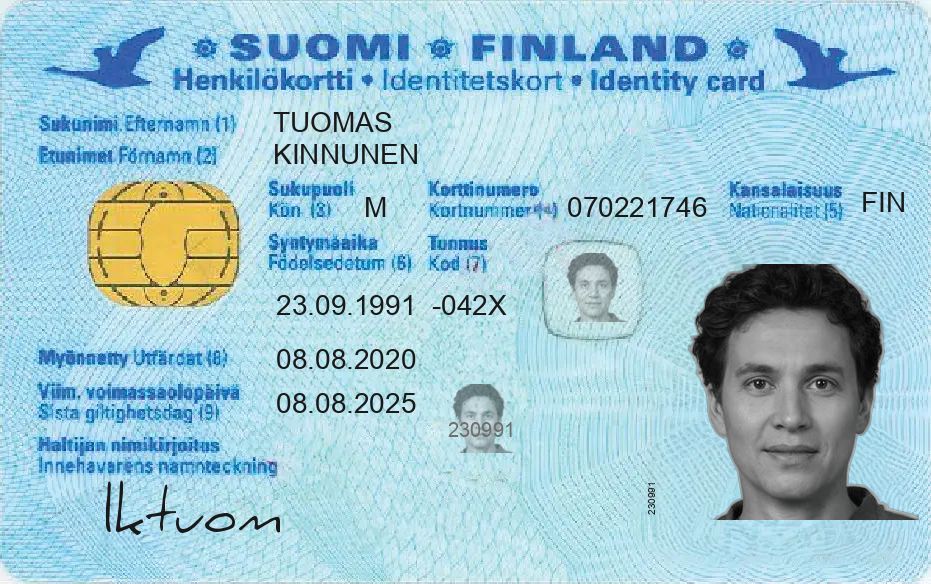}
\includegraphics[scale=0.13]{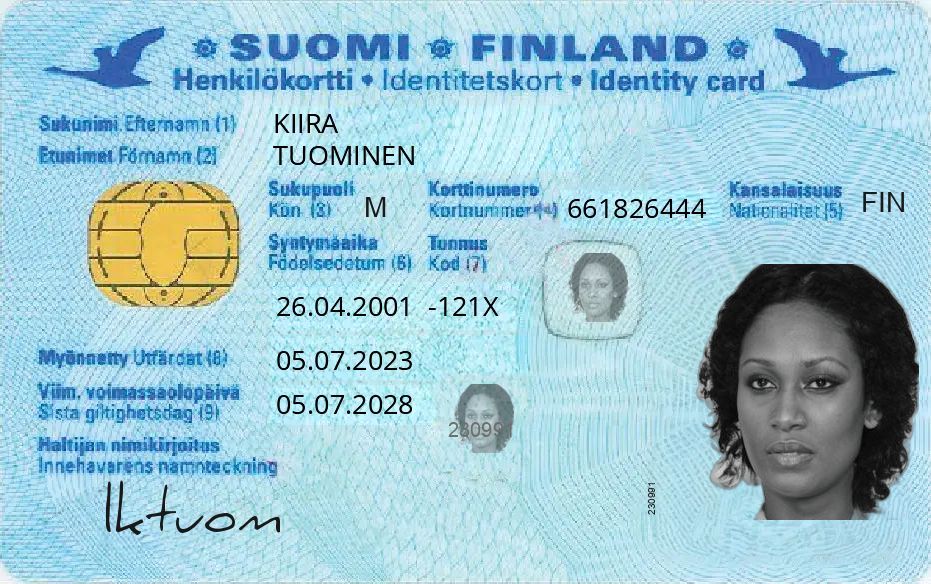}
\includegraphics[scale=0.13]{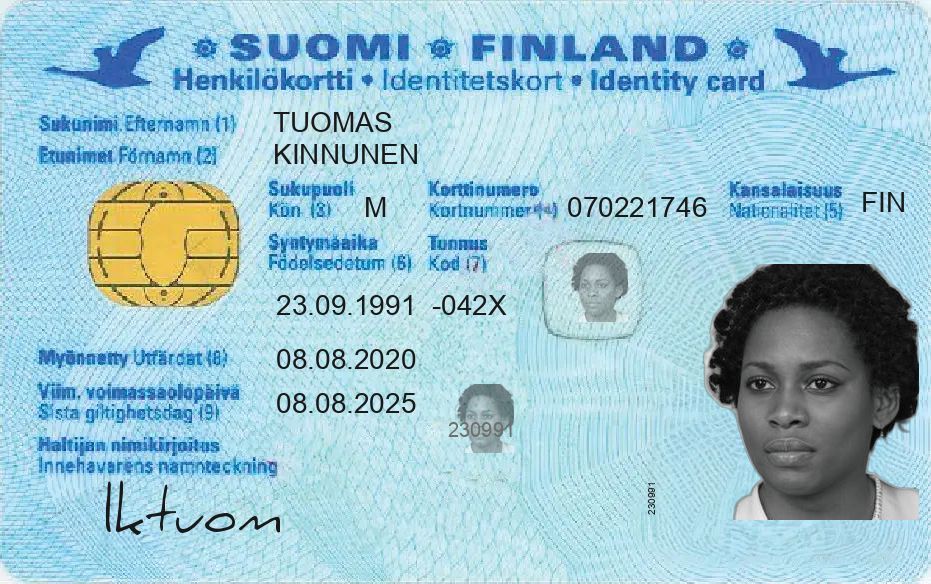}
\includegraphics[scale=0.13]{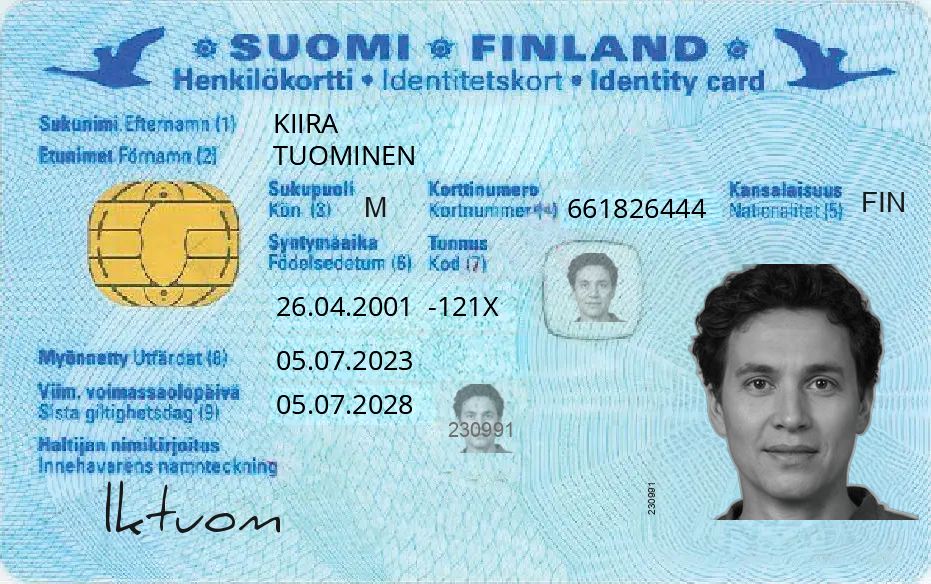}
\caption{\textbf{Top}: Example of images of the PAD Chilean ID cards CHL-DBP dataset. Left to right: Bona fide, Composite, Printed and Screen. \textbf{Bottom}: ID-Net example of open-set dataset (ID-Net). Left to right: Bona fide, combined, face-swap, copy-move attack.}
\label{fig:examples}
\end{figure*}

An example of the images is presented in Figure \ref{fig:examples}. Further on that, a summary table of all the images per class is shown in Table \ref{tab:my-tabledb}.

\begin{table}[]
\centering
\scriptsize
\caption{Summary Dataset Per Class. ID-Net is considering three countries: Finland, Spain, and Slovakia, with 5,980 images each.}
\label{tab:my-tabledb}
\begin{tabular}{ccccc}
\hline
\multicolumn{1}{c|}{Classes} &
  \multicolumn{1}{c|}{Train} &
  \multicolumn{1}{c|}{Test} &
  \multicolumn{1}{c|}{Validation} &
  \begin{tabular}[c]{@{}c@{}}Total \\ per class\end{tabular} \\ \hline
\multicolumn{5}{c}{Private Dataset (CHL-DBP)}                                                                                                                       \\ \hline
\multicolumn{1}{|c|}{Bona fide}     & \multicolumn{1}{c|}{3,510}  & \multicolumn{1}{c|}{1,171} & \multicolumn{1}{c|}{1,171} & \multicolumn{1}{c|}{5,852}  \\ \hline
\multicolumn{1}{|c|}{Border}        & \multicolumn{1}{c|}{13,397} & \multicolumn{1}{c|}{4,435} & \multicolumn{1}{c|}{4,430} & \multicolumn{1}{c|}{22,262} \\ \hline
\multicolumn{1}{|c|}{Printed}       & \multicolumn{1}{c|}{4,515}  & \multicolumn{1}{c|}{1,505} & \multicolumn{1}{c|}{1,508} & \multicolumn{1}{c|}{7,528}  \\ \hline
\multicolumn{1}{|c|}{Screen}        & \multicolumn{1}{c|}{3,180}  & \multicolumn{1}{c|}{1,062} & \multicolumn{1}{c|}{1,063} & \multicolumn{1}{c|}{5,305}  \\ \hline
\multicolumn{1}{|c|}{\begin{tabular}[c]{@{}c@{}}Total per\\ set\end{tabular}} &
  \multicolumn{1}{c|}{8,173} &
  \multicolumn{1}{c|}{24,602} &
  \multicolumn{1}{c|}{8,172} &
  \multicolumn{1}{c|}{\textbf{40,947}} \\ \hline
\multicolumn{5}{c}{IDNet}                                                                                                                                 \\ \hline
\multicolumn{1}{|c|}{Bona fide}     & \multicolumn{1}{c|}{10,000} & \multicolumn{1}{c|}{3,000} & \multicolumn{1}{c|}{4,940} & \multicolumn{1}{c|}{17,940} \\ \hline
\multicolumn{1}{|c|}{Combined}      & \multicolumn{1}{c|}{10,000} & \multicolumn{1}{c|}{3,000} & \multicolumn{1}{c|}{4,940} & \multicolumn{1}{c|}{17,940} \\ \hline
\multicolumn{1}{|c|}{Face-Swap}     & \multicolumn{1}{c|}{10,000} & \multicolumn{1}{c|}{3,000} & \multicolumn{1}{c|}{4,940} & \multicolumn{1}{c|}{17,940} \\ \hline
\multicolumn{1}{|c|}{Copy-Move}     & \multicolumn{1}{c|}{10,000} & \multicolumn{1}{c|}{3,000} & \multicolumn{1}{c|}{4,940} & \multicolumn{1}{c|}{17,940} \\ \hline
\multicolumn{1}{|c|}{Face-Morphing} & \multicolumn{1}{c|}{10,000} & \multicolumn{1}{c|}{3,000} & \multicolumn{1}{c|}{4,940} & \multicolumn{1}{c|}{17,940} \\ \hline
\multicolumn{1}{|c|}{\begin{tabular}[c]{@{}c@{}}Total per\\ set\end{tabular}} &
  \multicolumn{1}{c|}{50,000} &
  \multicolumn{1}{c|}{15,000} &
  \multicolumn{1}{c|}{24,700} &
  \multicolumn{1}{c|}{\textbf{89,700}} \\ \hline
\end{tabular}%
\end{table}

\section{Metric}
\label{sec:metrics}
The ISO/IEC 30107-3 standard\footnote{\url{https://www.iso.org/standard/67381.html}} presents methodologies for evaluating the performance of PAD algorithms for biometric systems. The APCER metric measures the proportion of attack presentations---for each different Presentation Attack Instrument (PAI)---incorrectly classified as bona fide presentations. This metric is calculated for each PAI, where the worst-case scenario is considered. Equation~\ref{eq:apcer} details how to compute the APCER metric, in which the value of $N_{PAIS}$ corresponds to the number of attack presentation images, where $RES_{i}$ for the $i$th image is $1$ if the algorithm classifies it as an attack presentation, or $0$ if it is classified as a bona fide presentation (real image).

\begin{equation}\label{eq:apcer}
    {APCER_{PAIS}}=1 - (\frac{1}{N_{PAIS}})\sum_{i=1}^{N_{PAIS}}RES_{i}
\end{equation}

Additionally, the BPCER metric measures the proportion of bona fide presentations mistakenly classified as attack presentations or the ratio between false rejection and total bona fide attempts. The BPCER metric is formulated according to equation~\ref{eq:bpcer}, where $N_{BF}$ corresponds to the number of bona fide presentation images, and $RES_{i}$ takes identical values to those of the APCER metric.

\begin{equation}\label{eq:bpcer}
    BPCER=\frac{\sum_{i=1}^{N_{BF}}RES_{i}}{N_{BF}}
\end{equation}

These metrics effectively measure to what degree the algorithm confuses presentations of attack images with bona fide images and vice versa. The APCER and BPCER metrics depend on a decision threshold.

\section{Experiment and Results}
\label{sec:results}

All the images were preprocessed based on the crop detection of the ID card using a retrained object detector and segmenter based on YoloV8 \cite{yolo}, but now, only ID cards and passport images are used. All the images were resized to the input of each network.

For the fine-tuning approach, a small set of data-augmentation functions was used, considering illumination changes, random rotation by 90 degrees, blurring and jitter.

All the experiments reported in this work are analysed considering the network training from scratch as a baseline, zero-shot, fine-tuning, and the Leave-One-Out (LOO) protocol to simulate unknown attacks.

\subsection{Experiment 1 - baseline}
In order to show the generalisation capabilities of deep learning models, we evaluated five classifiers trained from scratch based on ImageNet weights as a baseline on two classes bona fide and attacks. The networks selected were: DenseNet121 \cite{DenseNet}, EfficientNetV2-s \cite{efficientnet}, MobileNetV3-large \cite{mbv3}, ResNet34, and ResNet101 \cite{resnet}. Also, a Swin-Transformer was used based on Vision Transformer capabilities that make a difference with traditional deep learning models. The models evaluated are: SwinT-Base, SwinT-Small, and SwinT-large \cite{swin}. 
All the models were trained using SGD and Adam optimisers. Using a grid search, different learning rates were explored, ranging from $1e-3$ to $1e-6$. The $lr$ value of $1e-3$ achieved the best results.

For the CHL-DBP dataset, the best result was obtained by the SwinT-Small model with an EER of 5.45\%. 

In case of open-set ID-Net, all the models reached perfect results, less than 1,0\% of EER for all the models, which shows, according to our results based on the ID card subset, that this open-set dataset (ID-Net) in an intra-dataset evaluation is not very challenging. 

\subsection{Experiment 2 - Foundation models on zero-shot}

According to the analysis of the state-of-the-art, the DinoV2 and CLIP foundation models were not used in zero-shot modality with the LOO protocol to compare and analyse the generalisation to ID Card. 

For FM in zero-shot and LOO approaches, the pre-trained models were used to extract the features (embeddings). The size of each embedding for each deep learning model was set as follows: MobileNet, EfficientNet and ResNet were set to $1\times1280$. For DenseNet we choose $1\times1024$. For Vision Transformer models, SwinT was set to $1\times708$. 

For DinoV2-small $1\times384$, DinoV2-base $1\times708$ and DinoV2-large $1\times1024$. For CLIP, we have two variants with different patch sizes. CLIP-ViT-B16 and 32 divide images into $16\times16$ or $32\times32$ pixel patches. CLIP-ViT-L has more layers and parameters than the base model (B), enabling it to capture more complex features from the input data. This means that the model processes more patches with smaller dimensions, allowing it to focus on finer details. For CLIP-ViT-B and CLIP-ViT-L, the embedding size was set to $1\times256$ and $1\times708$, respectively. 

A single neuron is set up for the bona fide vs attack decision. At the same time, the weights of the model remained unchanged. 

Three different versions of DinoV2 were explored based on DinoV2-Base, DinoV2-Small and DinoV2-Large. 
Furthermore, for CLIP-ViT-L14, CLIP-ViT-B16 and CLIP-ViT-L32 were adapted. All the Clip models were explored based on the MetaClip-400-m databases. 

We explore the LOO protocol on the three different attacks available in the CHL-DBP dataset, print, screen, and border simulating and unknown attack was not used in the training and validation sets. For example, a bona fide vs printed and screen attack were used in the training and validation sets and the border attack was used as a test set. Then, bona fide vs border and screen were used in the training and validation sets and printed was used as a test set and so on. 

Table \ref{tab:zero-shot} shows the results for zero-shot in a LOO protocol in terms of EER, BPCER10, BPCER20 and BPCER100 for the CHL-DBP dataset. The models were used as a feature extractor, all the layers were frozen, and a sigmoid was used as a classifier. 

For Foundation models the best results were obtained by Dinov2-vitl14 with and EER of 4.33\%, followed for CLIP-ViT-B-16 with an EER of 13.31\%. According to the results, the printed attack is the most challenging.

For the deep learning models the best result was reached by ResNet101 in all three LOO modalities with an EER\% of 13.77 for border, 24.80\%  printed and 24.02\% screen as a test set, respectively. According to the results, the again printed attack is the most challenging.

For the Vision Transformer SwinTv2-s reached the best results with an EER of 12.63\%. The screen is the most challenging attack.


\begin{table*}[]
\caption{Summary result from \textbf{Zero-Shot} in the LOO protocol for Border, printed and screen attacks as unknown attack for CHL-DB.}
\label{tab:zero-shot}
\resizebox{\textwidth}{!}{%
\begin{tabular}{|ccccclccccccc|}
\hline
\multicolumn{1}{|c|}{Attack} &
  \multicolumn{4}{c|}{Border} &
  \multicolumn{4}{c|}{Printed} &
  \multicolumn{4}{c|}{Screen} \\ \hline
\multicolumn{1}{|c|}{Models} &
  \multicolumn{1}{c|}{\begin{tabular}[c]{@{}c@{}}EER\\ (\%)\end{tabular}} &
  \multicolumn{1}{c|}{\begin{tabular}[c]{@{}c@{}}BPCER10\\ (\%)\end{tabular}} &
  \multicolumn{1}{c|}{\begin{tabular}[c]{@{}c@{}}BPCER20\\ (\%)\end{tabular}} &
  \multicolumn{1}{c|}{\begin{tabular}[c]{@{}c@{}}BPCER100\\ (\%)\end{tabular}} &
  \multicolumn{1}{c|}{\begin{tabular}[c]{@{}c@{}}EER\\ (\%)\end{tabular}} &
  \multicolumn{1}{c|}{\begin{tabular}[c]{@{}c@{}}BPCER10\\ (\%)\end{tabular}} &
  \multicolumn{1}{c|}{\begin{tabular}[c]{@{}c@{}}BPCER20\\ (\%)\end{tabular}} &
  \multicolumn{1}{c|}{\begin{tabular}[c]{@{}c@{}}BPCER100\\ (\%)\end{tabular}} &
  \multicolumn{1}{c|}{\begin{tabular}[c]{@{}c@{}}EER\\ (\%)\end{tabular}} &
  \multicolumn{1}{c|}{\begin{tabular}[c]{@{}c@{}}BPCER10\\ (\%)\end{tabular}} &
  \multicolumn{1}{c|}{\begin{tabular}[c]{@{}c@{}}BPCER20\\ (\%)\end{tabular}} &
  \begin{tabular}[c]{@{}c@{}}BPCER100\\ (\%)\end{tabular} \\ \hline
\multicolumn{13}{|c|}{CLIP} \\ \hline
\multicolumn{1}{|c|}{CLIP\_ViT-B-16} &
  \multicolumn{1}{c|}{\mycc13.31} &
  \multicolumn{1}{c|}{\mycc17.25} &
  \multicolumn{1}{c|}{\mycc25.44} &
  \multicolumn{1}{c|}{\mycc46.02} &
  \multicolumn{1}{l|}{26.47} &
  \multicolumn{1}{c|}{59.94} &
  \multicolumn{1}{c|}{77.88} &
  \multicolumn{1}{c|}{93.76} &
  \multicolumn{1}{c|}{25.87} &
  \multicolumn{1}{c|}{58.83} &
  \multicolumn{1}{c|}{76.34} &
  92.99 \\ \hline
\multicolumn{1}{|c|}{CLIP\_ViT-B-32} &
  \multicolumn{1}{c|}{18.07} &
  \multicolumn{1}{c|}{27.41} &
  \multicolumn{1}{c|}{40.30} &
  \multicolumn{1}{c|}{66.52} &
  \multicolumn{1}{l|}{27.66} &
  \multicolumn{1}{c|}{60.03} &
  \multicolumn{1}{c|}{74.80} &
  \multicolumn{1}{c|}{91.97} &
  \multicolumn{1}{c|}{27.26} &
  \multicolumn{1}{c|}{58.07} &
  \multicolumn{1}{c|}{73.52} &
  91.97 \\ \hline
\multicolumn{1}{|c|}{CLIP\_ViT-L-14} &
  \multicolumn{1}{c|}{12.93} &
  \multicolumn{1}{c|}{15.88} &
  \multicolumn{1}{c|}{22.97} &
  \multicolumn{1}{c|}{47.56} &
  \multicolumn{1}{l|}{\mycc21.17} &
  \multicolumn{1}{c|}{\mycc44.83} &
  \multicolumn{1}{c|}{\mycc64.04} &
  \multicolumn{1}{c|}{\mycc85.99} &
  \multicolumn{1}{c|}{\mycc21.01} &
  \multicolumn{1}{c|}{\mycc44.83} &
  \multicolumn{1}{c|}{\mycc65.92} &
  \mycc87.27 \\ \hline
\multicolumn{13}{|c|}{Dino V2} \\ \hline
\multicolumn{1}{|c|}{Dinov2\_vitb14} &
  \multicolumn{1}{c|}{5.73} &
  \multicolumn{1}{c|}{2.90} &
  \multicolumn{1}{c|}{6.74} &
  \multicolumn{1}{c|}{26.98} &
  \multicolumn{1}{l|}{\mycc15.74} &
  \multicolumn{1}{c|}{\mycc27.32} &
  \multicolumn{1}{c|}{\mycc43.21} &
  \multicolumn{1}{c|}{\mycc76.17} &
  \multicolumn{1}{c|}{16.16} &
  \multicolumn{1}{c|}{28.69} &
  \multicolumn{1}{c|}{46.79} &
  79.33 \\ \hline
\multicolumn{1}{|c|}{Dinov2\_vits14} &
  \multicolumn{1}{c|}{7.79} &
  \multicolumn{1}{c|}{6.23} &
  \multicolumn{1}{c|}{15.02} &
  \multicolumn{1}{c|}{36.20} &
  \multicolumn{1}{l|}{17.94} &
  \multicolumn{1}{c|}{33.98} &
  \multicolumn{1}{c|}{52.69} &
  \multicolumn{1}{c|}{81.46} &
  \multicolumn{1}{c|}{16.74} &
  \multicolumn{1}{c|}{29.80} &
  \multicolumn{1}{c|}{53.03} &
  80.87 \\ \hline
\multicolumn{1}{|c|}{Dinov2\_vitl14} &
  \multicolumn{1}{c|}{\mycc\textbf{4.33}} &
  \multicolumn{1}{c|}{\mycc1.36} &
  \multicolumn{1}{c|}{\mycc3.41} &
  \multicolumn{1}{c|}{\mycc22.71} &
  \multicolumn{1}{l|}{15.82} &
  \multicolumn{1}{c|}{24.33} &
  \multicolumn{1}{c|}{40.64} &
  \multicolumn{1}{c|}{71.05} &
  \multicolumn{1}{c|}{\mycc15.19} &
  \multicolumn{1}{c|}{\mycc23.56} &
  \multicolumn{1}{c|}{\mycc40.22} &
  \mycc72.75 \\ \hline
\multicolumn{13}{|c|}{Deep Learning} \\ \hline
\multicolumn{1}{|c|}{DenseNet121} &
  \multicolumn{1}{c|}{15.91} &
  \multicolumn{1}{c|}{24.50} &
  \multicolumn{1}{c|}{36.54} &
  \multicolumn{1}{c|}{63.79} &
  \multicolumn{1}{l|}{25.54} &
  \multicolumn{1}{c|}{56.25} &
  \multicolumn{1}{c|}{72.84} &
  \multicolumn{1}{c|}{91.20} &
  \multicolumn{1}{c|}{24.62} &
  \multicolumn{1}{c|}{52.17} &
  \multicolumn{1}{c|}{66.78} &
  91.20 \\ \hline
\multicolumn{1}{|c|}{Efficientnet\_v2\_s} &
  \multicolumn{1}{c|}{22.40} &
  \multicolumn{1}{c|}{43.97} &
  \multicolumn{1}{c|}{58.75} &
  \multicolumn{1}{c|}{81.63} &
  \multicolumn{1}{l|}{29.05} &
  \multicolumn{1}{c|}{61.99} &
  \multicolumn{1}{c|}{77.11} &
  \multicolumn{1}{c|}{94.02} &
  \multicolumn{1}{c|}{29.32} &
  \multicolumn{1}{c|}{62.25} &
  \multicolumn{1}{c|}{76.17} &
  94.53 \\ \hline
\multicolumn{1}{|c|}{Mobilenet\_v3\_large} &
  \multicolumn{1}{c|}{20.47} &
  \multicolumn{1}{c|}{39.36} &
  \multicolumn{1}{c|}{55.76} &
  \multicolumn{1}{c|}{82.06} &
  \multicolumn{1}{l|}{27.02} &
  \multicolumn{1}{c|}{62.45} &
  \multicolumn{1}{c|}{76.60} &
  \multicolumn{1}{c|}{93.33} &
  \multicolumn{1}{c|}{25.90} &
  \multicolumn{1}{c|}{60.20} &
  \multicolumn{1}{c|}{78.05} &
  93.33 \\ \hline
\multicolumn{1}{|c|}{ResNet34} &
  \multicolumn{1}{c|}{17.74} &
  \multicolumn{1}{c|}{32.02} &
  \multicolumn{1}{c|}{48.59} &
  \multicolumn{1}{c|}{72.75} &
  \multicolumn{1}{l|}{26.05} &
  \multicolumn{1}{c|}{57.81} &
  \multicolumn{1}{c|}{74.03} &
  \multicolumn{1}{c|}{92.05} &
  \multicolumn{1}{c|}{25.80} &
  \multicolumn{1}{c|}{55.76} &
  \multicolumn{1}{c|}{74.72} &
  92.14 \\ \hline
\multicolumn{1}{|c|}{ResNet101} &
  \multicolumn{1}{c|}{\mycc13.77} &
  \multicolumn{1}{c|}{\mycc19.47} &
  \multicolumn{1}{c|}{\mycc38.81} &
  \multicolumn{1}{c|}{\mycc64.90} &
  \multicolumn{1}{l|}{\mycc24.80} &
  \multicolumn{1}{c|}{\mycc51.92} &
  \multicolumn{1}{c|}{\mycc71.39} &
  \multicolumn{1}{c|}{\mycc90.09} &
  \multicolumn{1}{c|}{\mycc24.02} &
  \multicolumn{1}{c|}{\mycc50.46} &
  \multicolumn{1}{c|}{\mycc68.40} &
  \mycc88.89 \\ \hline
\multicolumn{13}{|c|}{Vision Transformers} \\ \hline
\multicolumn{1}{|c|}{Swin\_v2\_b} &
  \multicolumn{1}{c|}{13.33} &
  \multicolumn{1}{c|}{16.05} &
  \multicolumn{1}{c|}{27.49} &
  \multicolumn{1}{c|}{59.35} &
  \multicolumn{1}{l|}{25.03} &
  \multicolumn{1}{c|}{53.62} &
  \multicolumn{1}{c|}{69.00} &
  \multicolumn{1}{c|}{86.84} &
  \multicolumn{1}{c|}{24.07} &
  \multicolumn{1}{c|}{54.31} &
  \multicolumn{1}{c|}{66.74} &
  85.55 \\ \hline
\multicolumn{1}{|c|}{Swin\_v2\_s} &
  \multicolumn{1}{c|}{\mycc12.63} &
  \multicolumn{1}{c|}{\mycc15.45} &
  \multicolumn{1}{c|}{\mycc29.20} &
  \multicolumn{1}{c|}{\mycc54.48} &
  \multicolumn{1}{l|}{24.09} &
  \multicolumn{1}{c|}{49.70} &
  \multicolumn{1}{c|}{65.15} &
  \multicolumn{1}{c|}{90.60} &
  \multicolumn{1}{c|}{21.97} &
  \multicolumn{1}{c|}{45.94} &
  \multicolumn{1}{c|}{65.24} &
  88.12 \\ \hline
\multicolumn{1}{|c|}{Swin\_v2\_l} &
  \multicolumn{1}{c|}{12.62} &
  \multicolumn{1}{c|}{16.14} &
  \multicolumn{1}{c|}{30.31} &
  \multicolumn{1}{c|}{64.21} &
  \multicolumn{1}{l|}{\mycc22.79} &
  \multicolumn{1}{c|}{\mycc44.06} &
  \multicolumn{1}{c|}{\mycc60.80} &
  \multicolumn{1}{c|}{\mycc84.97} &
  \multicolumn{1}{c|}{\mycc21.40} &
  \multicolumn{1}{c|}{\mycc43.72} &
  \multicolumn{1}{c|}{\mycc60.63} &
  \mycc85.39 \\ \hline
\end{tabular}%
}
\end{table*}

\begin{table*}[]
\caption{Summary results from \textbf{Fine-tuning} in the LOO protocol considering border, printed and screen as an unknown attack for \textbf{CHL-DBP}.}
\label{tab:finetuning-loo}
\resizebox{\textwidth}{!}{%
\begin{tabular}{|ccccclccccccc|}
\hline
\multicolumn{1}{|c|}{Attack} &
  \multicolumn{4}{c|}{Border} &
  \multicolumn{4}{c|}{Printed} &
  \multicolumn{4}{c|}{Screen} \\ \hline
\multicolumn{1}{|c|}{Models} &
  \multicolumn{1}{c|}{\begin{tabular}[c]{@{}c@{}}EER\\ (\%)\end{tabular}} &
  \multicolumn{1}{c|}{\begin{tabular}[c]{@{}c@{}}BPCER10\\ (\%)\end{tabular}} &
  \multicolumn{1}{c|}{\begin{tabular}[c]{@{}c@{}}BPCER20\\ (\%)\end{tabular}} &
  \multicolumn{1}{c|}{\begin{tabular}[c]{@{}c@{}}BPCER100\\ (\%)\end{tabular}} &
  \multicolumn{1}{c|}{\begin{tabular}[c]{@{}c@{}}EER\\ (\%)\end{tabular}} &
  \multicolumn{1}{c|}{\begin{tabular}[c]{@{}c@{}}BPCER10\\ (\%)\end{tabular}} &
  \multicolumn{1}{c|}{\begin{tabular}[c]{@{}c@{}}BPCER20\\ (\%)\end{tabular}} &
  \multicolumn{1}{c|}{\begin{tabular}[c]{@{}c@{}}BPCER100\\ (\%)\end{tabular}} &
  \multicolumn{1}{c|}{\begin{tabular}[c]{@{}c@{}}EER\\ (\%)\end{tabular}} &
  \multicolumn{1}{c|}{\begin{tabular}[c]{@{}c@{}}BPCER10\\ (\%)\end{tabular}} &
  \multicolumn{1}{c|}{\begin{tabular}[c]{@{}c@{}}BPCER20\\ (\%)\end{tabular}} &
  \begin{tabular}[c]{@{}c@{}}BPCER100\\ (\%)\end{tabular} \\ \hline
\multicolumn{13}{|c|}{CLIP} \\ \hline
\multicolumn{1}{|c|}{CLIP\_ViT-B-16} &
  \multicolumn{1}{c|}{12.23} &
  \multicolumn{1}{c|}{14.60} &
  \multicolumn{1}{c|}{23.22} &
  \multicolumn{1}{c|}{43.80} &
  \multicolumn{1}{l|}{27.14} &
  \multicolumn{1}{c|}{62.16} &
  \multicolumn{1}{c|}{77.36} &
  \multicolumn{1}{c|}{94.02} &
  \multicolumn{1}{c|}{26.53} &
  \multicolumn{1}{c|}{59.86} &
  \multicolumn{1}{c|}{76.51} &
  94.53 \\ \hline
\multicolumn{1}{|c|}{CLIP\_ViT-B-32} &
  \multicolumn{1}{c|}{18.01} &
  \multicolumn{1}{c|}{26.43} &
  \multicolumn{1}{c|}{39.02} &
  \multicolumn{1}{c|}{69.00} &
  \multicolumn{1}{l|}{29.71} &
  \multicolumn{1}{c|}{64.13} &
  \multicolumn{1}{c|}{64.13} &
  \multicolumn{1}{c|}{92.91} &
  \multicolumn{1}{c|}{28.60} &
  \multicolumn{1}{c|}{62.08} &
  \multicolumn{1}{c|}{74.97} &
  93.25 \\ \hline
\multicolumn{1}{|c|}{CLIP\_ViT-L-14} &
  \multicolumn{1}{c|}{\mycc11.52} &
  \multicolumn{1}{c|}{\mycc12.55} &
  \multicolumn{1}{c|}{\mycc16.64} &
  \multicolumn{1}{c|}{\mycc44.57} &
  \multicolumn{1}{l|}{\mycc19.97} &
  \multicolumn{1}{c|}{\mycc41.50} &
  \multicolumn{1}{c|}{\mycc60.80} &
  \multicolumn{1}{c|}{\mycc84.97} &
  \multicolumn{1}{c|}{\mycc20.01} &
  \multicolumn{1}{c|}{\mycc42.35} &
  \multicolumn{1}{c|}{\mycc63.02} &
  \mycc87.01 \\ \hline
\multicolumn{13}{|c|}{Dino V2} \\ \hline
\multicolumn{1}{|c|}{Dinov2\_vitb14} &
  \multicolumn{1}{c|}{5.57} &
  \multicolumn{1}{c|}{2.90} &
  \multicolumn{1}{c|}{6.40} &
  \multicolumn{1}{c|}{22.80} &
  \multicolumn{1}{l|}{\mycc14.97} &
  \multicolumn{1}{c|}{\mycc23.99} &
  \multicolumn{1}{c|}{\mycc42.18} &
  \multicolumn{1}{c|}{\mycc75.40} &
  \multicolumn{1}{c|}{15.36} &
  \multicolumn{1}{c|}{25.96} &
  \multicolumn{1}{c|}{44.23} &
  75.57 \\ \hline
\multicolumn{1}{|c|}{Dinov2\_vits14} &
  \multicolumn{1}{c|}{\mycc\textbf{4.13}} &
  \multicolumn{1}{c|}{\mycc1.28} &
  \multicolumn{1}{c|}{\mycc2.73} &
  \multicolumn{1}{c|}{\mycc19.72} &
  \multicolumn{1}{l|}{15.19} &
  \multicolumn{1}{c|}{23.39} &
  \multicolumn{1}{c|}{38.25} &
  \multicolumn{1}{c|}{69.59} &
  \multicolumn{1}{c|}{15.89} &
  \multicolumn{1}{c|}{26.64} &
  \multicolumn{1}{c|}{46.11} &
  75.49 \\ \hline
\multicolumn{1}{|c|}{Dinov2\_vitl14} &
  \multicolumn{1}{c|}{23.73} &
  \multicolumn{1}{c|}{40.22} &
  \multicolumn{1}{c|}{52.69} &
  \multicolumn{1}{c|}{71.39} &
  \multicolumn{1}{l|}{17.03} &
  \multicolumn{1}{c|}{28.26} &
  \multicolumn{1}{c|}{44.83} &
  \multicolumn{1}{c|}{76.08} &
  \multicolumn{1}{c|}{\mycc14.78} &
  \multicolumn{1}{c|}{\mycc22.45} &
  \multicolumn{1}{c|}{\mycc36.72} &
  \mycc70.67 \\ \hline
\multicolumn{13}{|c|}{Deep Learning} \\ \hline
\multicolumn{1}{|c|}{DenseNet121} &
  \multicolumn{1}{c|}{14.71} &
  \multicolumn{1}{c|}{23.25} &
  \multicolumn{1}{c|}{34.50} &
  \multicolumn{1}{c|}{61.19} &
  \multicolumn{1}{l|}{23.50} &
  \multicolumn{1}{c|}{53.15} &
  \multicolumn{1}{c|}{71.04} &
  \multicolumn{1}{c|}{89.10} &
  \multicolumn{1}{c|}{23.05} &
  \multicolumn{1}{c|}{50.17} &
  \multicolumn{1}{c|}{64.18} &
  91.20 \\ \hline
\multicolumn{1}{|c|}{Efficientnet\_v2\_s} &
  \multicolumn{1}{c|}{21.05} &
  \multicolumn{1}{c|}{41.87} &
  \multicolumn{1}{c|}{57.35} &
  \multicolumn{1}{c|}{80.03} &
  \multicolumn{1}{l|}{27.95} &
  \multicolumn{1}{c|}{59.35} &
  \multicolumn{1}{c|}{75.05} &
  \multicolumn{1}{c|}{92.12} &
  \multicolumn{1}{c|}{28.22} &
  \multicolumn{1}{c|}{61.05} &
  \multicolumn{1}{c|}{75.07} &
  94.53 \\ \hline
\multicolumn{1}{|c|}{Mobilenet\_v3\_large} &
  \multicolumn{1}{c|}{19.37} &
  \multicolumn{1}{c|}{38.06} &
  \multicolumn{1}{c|}{54.16} &
  \multicolumn{1}{c|}{80.05} &
  \multicolumn{1}{l|}{26.15} &
  \multicolumn{1}{c|}{60.95} &
  \multicolumn{1}{c|}{75.05} &
  \multicolumn{1}{c|}{91.35} &
  \multicolumn{1}{c|}{23.80} &
  \multicolumn{1}{c|}{58.75} &
  \multicolumn{1}{c|}{76.15} &
  89.33 \\ \hline
\multicolumn{1}{|c|}{ResNet34} &
  \multicolumn{1}{c|}{16.34} &
  \multicolumn{1}{c|}{30.95} &
  \multicolumn{1}{c|}{47.05} &
  \multicolumn{1}{c|}{71.15} &
  \multicolumn{1}{l|}{25.15} &
  \multicolumn{1}{c|}{55.33} &
  \multicolumn{1}{c|}{72.25} &
  \multicolumn{1}{c|}{89.15} &
  \multicolumn{1}{c|}{23.60} &
  \multicolumn{1}{c|}{54.36} &
  \multicolumn{1}{c|}{72.12} &
  99.07 \\ \hline
\multicolumn{1}{|c|}{ResNet101} &
  \multicolumn{1}{c|}{\mycc11.70} &
  \multicolumn{1}{c|}{\mycc18.35} &
  \multicolumn{1}{c|}{\mycc37.61} &
  \multicolumn{1}{c|}{\mycc62.85} &
  \multicolumn{1}{l|}{\mycc22.15} &
  \multicolumn{1}{c|}{\mycc50.25} &
  \multicolumn{1}{c|}{\mycc68.19} &
  \multicolumn{1}{c|}{\mycc86.33} &
  \multicolumn{1}{c|}{\mycc23.19} &
  \multicolumn{1}{c|}{\mycc49.16} &
  \multicolumn{1}{c|}{\mycc67.35} &
  \mycc85.39 \\ \hline
\multicolumn{13}{|c|}{Vision Transformers} \\ \hline
\multicolumn{1}{|c|}{Swin\_v2\_b\_} &
  \multicolumn{1}{c|}{11.31} &
  \multicolumn{1}{c|}{14.15} &
  \multicolumn{1}{c|}{25.95} &
  \multicolumn{1}{c|}{57.16} &
  \multicolumn{1}{l|}{24.35} &
  \multicolumn{1}{c|}{51.25} &
  \multicolumn{1}{c|}{67.15} &
  \multicolumn{1}{c|}{84.33} &
  \multicolumn{1}{c|}{21.17} &
  \multicolumn{1}{c|}{53.16} &
  \multicolumn{1}{c|}{64.14} &
  82.15 \\ \hline
\multicolumn{1}{|c|}{Swin\_v2\_s} &
  \multicolumn{1}{c|}{11.00} &
  \multicolumn{1}{c|}{13.15} &
  \multicolumn{1}{c|}{27.20} &
  \multicolumn{1}{c|}{49.18} &
  \multicolumn{1}{l|}{21.15} &
  \multicolumn{1}{c|}{47.33} &
  \multicolumn{1}{c|}{63.18} &
  \multicolumn{1}{c|}{87.10} &
  \multicolumn{1}{c|}{20.13} &
  \multicolumn{1}{c|}{43.00} &
  \multicolumn{1}{c|}{63.40} &
  85.25 \\ \hline
\multicolumn{1}{|c|}{Swin\_v2\_t} &
  \multicolumn{1}{c|}{\mycc10.02} &
  \multicolumn{1}{c|}{\mycc13.25} &
  \multicolumn{1}{c|}{\mycc27.23} &
  \multicolumn{1}{c|}{\mycc61.70} &
  \multicolumn{1}{l|}{\mycc20.03} &
  \multicolumn{1}{c|}{\mycc42.15} &
  \multicolumn{1}{c|}{\mycc55.00} &
  \multicolumn{1}{c|}{\mycc80.33} &
  \multicolumn{1}{c|}{\mycc18.30} &
  \multicolumn{1}{c|}{\mycc39.72} &
  \multicolumn{1}{c|}{\mycc58.13} &
  \mycc82.90 \\ \hline
\end{tabular}%
}
\end{table*}


\begin{table*}[]
\caption{Summary results from \textbf{Fine-tuning} in the LOO protocol considering border, printed and screen as an unknown attack for \textbf{ID-Net}.}
\label{tab:LOO-IDnet}
\resizebox{\textwidth}{!}{%
\begin{tabular}{|ccccccccccccc|}
\hline
\multicolumn{1}{|c|}{\multirow{2}{*}{Models}} &
  \multicolumn{1}{c|}{\begin{tabular}[c]{@{}c@{}}EER\\ (\%)\end{tabular}} &
  \multicolumn{1}{c|}{\begin{tabular}[c]{@{}c@{}}BPCER10\\ (\%)\end{tabular}} &
  \multicolumn{1}{c|}{\begin{tabular}[c]{@{}c@{}}BPCER20\\ (\%)\end{tabular}} &
  \multicolumn{1}{c|}{\begin{tabular}[c]{@{}c@{}}EER\\ (\%)\end{tabular}} &
  \multicolumn{1}{c|}{\begin{tabular}[c]{@{}c@{}}BPCER10\\ (\%)\end{tabular}} &
  \multicolumn{1}{c|}{\begin{tabular}[c]{@{}c@{}}BPCER20\\ (\%)\end{tabular}} &
  \multicolumn{1}{c|}{\begin{tabular}[c]{@{}c@{}}EER\\ (\%)\end{tabular}} &
  \multicolumn{1}{c|}{\begin{tabular}[c]{@{}c@{}}BPCER10\\ (\%)\end{tabular}} &
  \multicolumn{1}{c|}{\begin{tabular}[c]{@{}c@{}}BPCER20\\ (\%)\end{tabular}} &
  \multicolumn{1}{c|}{\begin{tabular}[c]{@{}c@{}}EER\\ (\%)\end{tabular}} &
  \multicolumn{1}{c|}{\begin{tabular}[c]{@{}c@{}}BPCER10\\ (\%)\end{tabular}} &
  \begin{tabular}[c]{@{}c@{}}BPCER20\\ (\%)\end{tabular} \\ \cline{2-13} 
\multicolumn{1}{|c|}{} &
  \multicolumn{3}{c|}{Face-replacement} &
  \multicolumn{3}{c|}{Face combined} &
  \multicolumn{3}{c|}{Face-morphing} &
  \multicolumn{3}{c|}{Face Copy-Move} \\ \hline
\multicolumn{13}{|c|}{CLIP} \\ \hline
\multicolumn{1}{|c|}{CLIP\_ViT-B-16} &
  \multicolumn{1}{c|}{22.95} &
  \multicolumn{1}{c|}{48.97} &
  \multicolumn{1}{c|}{66.22} &
  \multicolumn{1}{c|}{26.22} &
  \multicolumn{1}{c|}{53.71} &
  \multicolumn{1}{c|}{68.16} &
  \multicolumn{1}{c|}{15.32} &
  \multicolumn{1}{c|}{24.28} &
  \multicolumn{1}{c|}{39.48} &
  \multicolumn{1}{c|}{24.38} &
  \multicolumn{1}{c|}{52.65} &
  64.46 \\ \hline
\multicolumn{1}{|c|}{CLIP\_ViT-B-32} &
  \multicolumn{1}{c|}{\mycc21.36} &
  \multicolumn{1}{c|}{\mycc49.59} &
  \multicolumn{1}{c|}{\mycc66.62} &
  \multicolumn{1}{c|}{\mycc25.10} &
  \multicolumn{1}{c|}{\mycc55.61} &
  \multicolumn{1}{c|}{\mycc70.51} &
  \multicolumn{1}{c|}{\mycc12.48} &
  \multicolumn{1}{c|}{\mycc16.63} &
  \multicolumn{1}{c|}{\mycc33.46} &
  \multicolumn{1}{c|}{\mycc23.89} &
  \multicolumn{1}{c|}{\mycc54.08} &
  \mycc70.51 \\ \hline
\multicolumn{1}{|c|}{CLIP\_ViT-L-14} &
  \multicolumn{1}{c|}{26.02} &
  \multicolumn{1}{c|}{57.14} &
  \multicolumn{1}{c|}{72.85} &
  \multicolumn{1}{c|}{27.65} &
  \multicolumn{1}{c|}{59.28} &
  \multicolumn{1}{c|}{74.28} &
  \multicolumn{1}{c|}{16.25} &
  \multicolumn{1}{c|}{30.10} &
  \multicolumn{1}{c|}{46.32} &
  \multicolumn{1}{c|}{25.81} &
  \multicolumn{1}{c|}{57.14} &
  72.85 \\ \hline
\multicolumn{13}{|c|}{DinoV2} \\ \hline
\multicolumn{1}{|c|}{Dinov2\_vitb14} &
  \multicolumn{1}{c|}{12.85} &
  \multicolumn{1}{c|}{17.95} &
  \multicolumn{1}{c|}{34.48} &
  \multicolumn{1}{c|}{15.13} &
  \multicolumn{1}{c|}{24.48} &
  \multicolumn{1}{c|}{40.51} &
  \multicolumn{1}{c|}{6.03} &
  \multicolumn{1}{c|}{1.83} &
  \multicolumn{1}{c|}{8.36} &
  \multicolumn{1}{c|}{15.0} &
  \multicolumn{1}{c|}{22.55} &
  38.77 \\ \hline
\multicolumn{1}{|c|}{Dinov2\_vitl14} &
  \multicolumn{1}{c|}{\mycc11.73} &
  \multicolumn{1}{c|}{\mycc14.08} &
  \multicolumn{1}{c|}{\mycc27.75} &
  \multicolumn{1}{c|}{\mycc13.57} &
  \multicolumn{1}{c|}{\mycc19.18} &
  \multicolumn{1}{c|}{\mycc33.97} &
  \multicolumn{1}{c|}{\mycc\textbf{5.91}} &
  \multicolumn{1}{c|}{\mycc3.16} &
  \multicolumn{1}{c|}{\mycc6.93} &
  \multicolumn{1}{c|}{\mycc12.36} &
  \multicolumn{1}{c|}{\mycc17.55} &
  \mycc30.30 \\ \hline
\multicolumn{1}{|c|}{Dinov2\_vits14} &
  \multicolumn{1}{c|}{16.73} &
  \multicolumn{1}{c|}{29.28} &
  \multicolumn{1}{c|}{48.46} &
  \multicolumn{1}{c|}{19.89} &
  \multicolumn{1}{c|}{37.34} &
  \multicolumn{1}{c|}{53.36} &
  \multicolumn{1}{c|}{9.59} &
  \multicolumn{1}{c|}{8.57} &
  \multicolumn{1}{c|}{19.59} &
  \multicolumn{1}{c|}{18.40} &
  \multicolumn{1}{c|}{33.16} &
  50.81 \\ \hline
\multicolumn{13}{|c|}{Deep Learning} \\ \hline
\multicolumn{1}{|c|}{Densenet121} &
  \multicolumn{1}{c|}{20.81} &
  \multicolumn{1}{c|}{43.97} &
  \multicolumn{1}{c|}{62.55} &
  \multicolumn{1}{c|}{23.26} &
  \multicolumn{1}{c|}{45.81} &
  \multicolumn{1}{c|}{62.34} &
  \multicolumn{1}{c|}{13.09} &
  \multicolumn{1}{c|}{19.48} &
  \multicolumn{1}{c|}{32.65} &
  \multicolumn{1}{c|}{23.26} &
  \multicolumn{1}{c|}{47.14} &
  64.89 \\ \hline
\multicolumn{1}{|c|}{Efficientnet\_v2\_s} &
  \multicolumn{1}{c|}{25.42} &
  \multicolumn{1}{c|}{62.44} &
  \multicolumn{1}{c|}{79.08} &
  \multicolumn{1}{c|}{31.22} &
  \multicolumn{1}{c|}{68.67} &
  \multicolumn{1}{c|}{81.53} &
  \multicolumn{1}{c|}{19.01} &
  \multicolumn{1}{c|}{41.83} &
  \multicolumn{1}{c|}{59.38} &
  \multicolumn{1}{c|}{30.85} &
  \multicolumn{1}{c|}{67.95} &
  82.44 \\ \hline
\multicolumn{1}{|c|}{Mobilenet\_v3\_large} &
  \multicolumn{1}{c|}{17.67} &
  \multicolumn{1}{c|}{38.77} &
  \multicolumn{1}{c|}{57.95} &
  \multicolumn{1}{c|}{20.81} &
  \multicolumn{1}{c|}{41.73} &
  \multicolumn{1}{c|}{58.36} &
  \multicolumn{1}{c|}{11.12} &
  \multicolumn{1}{c|}{13.06} &
  \multicolumn{1}{c|}{29.28} &
  \multicolumn{1}{c|}{21.53} &
  \multicolumn{1}{c|}{43.36} &
  59.38 \\ \hline
\multicolumn{1}{|c|}{ResNet34} &
  \multicolumn{1}{c|}{22.75} &
  \multicolumn{1}{c|}{53.26} &
  \multicolumn{1}{c|}{70.91} &
  \multicolumn{1}{c|}{27.56} &
  \multicolumn{1}{c|}{59.59} &
  \multicolumn{1}{c|}{74.08} &
  \multicolumn{1}{c|}{15.93} &
  \multicolumn{1}{c|}{30.61} &
  \multicolumn{1}{c|}{47.04} &
  \multicolumn{1}{c|}{27.55} &
  \multicolumn{1}{c|}{61.22} &
  75.30 \\ \hline
\multicolumn{1}{|c|}{ResNet101} &
  \multicolumn{1}{c|}{\mycc16.54} &
  \multicolumn{1}{c|}{\mycc29.59} &
  \multicolumn{1}{c|}{\mycc47.14} &
  \multicolumn{1}{c|}{\mycc19.89} &
  \multicolumn{1}{c|}{\mycc35.51} &
  \multicolumn{1}{c|}{\mycc52.75} &
  \multicolumn{1}{c|}{\mycc10.81} &
  \multicolumn{1}{c|}{\mycc12.34} &
  \multicolumn{1}{c|}{\mycc24.18} &
  \multicolumn{1}{c|}{\mycc20.30} &
  \multicolumn{1}{c|}{\mycc37.85} &
  \mycc53.36 \\ \hline
\multicolumn{13}{|c|}{Vision Transformers} \\ \hline
\multicolumn{1}{|c|}{Swin\_v2\_b\_} &
  \multicolumn{1}{c|}{25.00} &
  \multicolumn{1}{c|}{62.75} &
  \multicolumn{1}{c|}{80.10} &
  \multicolumn{1}{c|}{28.97} &
  \multicolumn{1}{c|}{68.16} &
  \multicolumn{1}{c|}{83.87} &
  \multicolumn{1}{c|}{27.95} &
  \multicolumn{1}{c|}{67.04} &
  \multicolumn{1}{c|}{82.44} &
  \multicolumn{1}{c|}{27.95} &
  \multicolumn{1}{c|}{67.04} &
  82.44 \\ \hline
\multicolumn{1}{|c|}{Swin\_v2\_s} &
  \multicolumn{1}{c|}{24.71} &
  \multicolumn{1}{c|}{61.02} &
  \multicolumn{1}{c|}{77.95} &
  \multicolumn{1}{c|}{28.50} &
  \multicolumn{1}{c|}{64.59} &
  \multicolumn{1}{c|}{79.28} &
  \multicolumn{1}{c|}{14.69} &
  \multicolumn{1}{c|}{22.24} &
  \multicolumn{1}{c|}{38.77} &
  \multicolumn{1}{c|}{28.06} &
  \multicolumn{1}{c|}{65.51} &
  80.20 \\ \hline
\multicolumn{1}{|c|}{Swin\_v2\_t} &
  \multicolumn{1}{c|}{\mycc24.69} &
  \multicolumn{1}{c|}{\mycc57.14} &
  \multicolumn{1}{c|}{\mycc75.40} &
  \multicolumn{1}{c|}{\mycc26.95} &
  \multicolumn{1}{c|}{\mycc60.40} &
  \multicolumn{1}{c|}{\mycc76.73} &
  \multicolumn{1}{c|}{\mycc13.46} &
  \multicolumn{1}{c|}{\mycc20.71} &
  \multicolumn{1}{c|}{\mycc36.83} &
  \multicolumn{1}{c|}{\mycc26.02} &
  \multicolumn{1}{c|}{\mycc61.22} &
  \mycc77.44 \\ \hline
\end{tabular}%
}
\end{table*}

\begin{table*}[]
\centering
\scriptsize
\caption{Summary Results model trained from \textbf{fine-tuning} considering \textbf{two classes} for the CHL-DBP dataset and open-set ID-Net.}
\label{tabla2c}
\begin{tabular}{|ccccccccc|}
\hline
\multicolumn{1}{|c|}{Models} &
  \multicolumn{1}{c|}{\begin{tabular}[c]{@{}c@{}}EER\\ (\%)\end{tabular}} &
  \multicolumn{1}{c|}{\begin{tabular}[c]{@{}c@{}}BPCER10\\ (\%)\end{tabular}} &
  \multicolumn{1}{c|}{\begin{tabular}[c]{@{}c@{}}BPCER20\\ (\%)\end{tabular}} &
  \multicolumn{1}{c|}{\begin{tabular}[c]{@{}c@{}}BPCER100\\ (\%)\end{tabular}} &
  \multicolumn{1}{c|}{\begin{tabular}[c]{@{}c@{}}EER\\ (\%)\end{tabular}} &
  \multicolumn{1}{c|}{\begin{tabular}[c]{@{}c@{}}BPCER10\\ (\%)\end{tabular}} &
  \multicolumn{1}{c|}{\begin{tabular}[c]{@{}c@{}}BPCER20\\ (\%)\end{tabular}} &
  \begin{tabular}[c]{@{}c@{}}BPCER100\\ (\%)\end{tabular} \\ \hline
\multicolumn{5}{|c|}{Private Dataset 
(CHL-DBP)} &
  \multicolumn{4}{c|}{Open-set (ID-Net)} \\ \hline
\multicolumn{9}{|c|}{CLIP} \\ \hline
\multicolumn{1}{|c|}{CLIP\_ViT-B-16} &
  \multicolumn{1}{c|}{11.95} &
  \multicolumn{1}{c|}{15.96} &
  \multicolumn{1}{c|}{31.16} &
  \multicolumn{1}{c|}{64.90} &
  \multicolumn{1}{c|}{41.33} &
  \multicolumn{1}{c|}{80.93} &
  \multicolumn{1}{c|}{89.46} &
  97.59 \\ \hline
\multicolumn{1}{|c|}{CLIP\_ViT-B-32} &
  \multicolumn{1}{c|}{14.01} &
  \multicolumn{1}{c|}{21.17} &
  \multicolumn{1}{c|}{37.91} &
  \multicolumn{1}{c|}{73.78} &
  \multicolumn{1}{c|}{40.95} &
  \multicolumn{1}{c|}{79.80} &
  \multicolumn{1}{c|}{88.53} &
  97.00 \\ \hline
\multicolumn{1}{|c|}{CLIP\_ViT-L-14} &
  \multicolumn{1}{c|}{\mycc9.51} &
  \multicolumn{1}{c|}{\mycc8.79} &
  \multicolumn{1}{c|}{\mycc22.88} &
  \multicolumn{1}{c|}{\mycc58.58} &
  \multicolumn{1}{c|}{\mycc40.86} &
  \multicolumn{1}{c|}{\mycc79.93} &
  \multicolumn{1}{c|}{\mycc80.80} &
  \mycc97.19 \\ \hline
\multicolumn{9}{|c|}{DinoV2} \\ \hline
\multicolumn{1}{|c|}{Dinov2\_vitb14} &
  \multicolumn{1}{c|}{6.48} &
  \multicolumn{1}{c|}{3.07} &
  \multicolumn{1}{c|}{9.05} &
  \multicolumn{1}{c|}{37.74} &
  \multicolumn{1}{c|}{\mycc\textbf{27.86}} &
  \multicolumn{1}{c|}{\mycc61.40} &
  \multicolumn{1}{c|}{\mycc76.13} &
  \mycc93.46 \\ \hline
\multicolumn{1}{|c|}{Dinov2\_vitl14} &
  \multicolumn{1}{c|}{\mycc\textbf{6.33}} &
  \multicolumn{1}{c|}{\mycc3.24} &
  \multicolumn{1}{c|}{\mycc9.13} &
  \multicolumn{1}{c|}{\mycc39.28} &
  \multicolumn{1}{c|}{32.67} &
  \multicolumn{1}{c|}{70.46} &
  \multicolumn{1}{c|}{84.13} &
  96.19 \\ \hline
\multicolumn{1}{|c|}{Dinov2\_vits14} &
  \multicolumn{1}{c|}{7.72} &
  \multicolumn{1}{c|}{5.46} &
  \multicolumn{1}{c|}{15.28} &
  \multicolumn{1}{c|}{55.42} &
  \multicolumn{1}{c|}{36.79} &
  \multicolumn{1}{c|}{74.13} &
  \multicolumn{1}{c|}{85.33} &
  95.80 \\ \hline
\multicolumn{9}{|c|}{Deep Learning} \\ \hline
\multicolumn{1}{|c|}{DenseNet121} &
  \multicolumn{1}{c|}{\mycc15.65} &
  \multicolumn{1}{c|}{\mycc26.21} &
  \multicolumn{1}{c|}{\mycc42.52} &
  \multicolumn{1}{c|}{\mycc74.12} &
  \multicolumn{1}{c|}{37.24} &
  \multicolumn{1}{c|}{74.69} &
  \multicolumn{1}{c|}{85.91} &
  97.65 \\ \hline
\multicolumn{1}{|c|}{Efficientnet\_v2\_s} &
  \multicolumn{1}{c|}{23.14} &
  \multicolumn{1}{c|}{47.73} &
  \multicolumn{1}{c|}{63.79} &
  \multicolumn{1}{c|}{86.93} &
  \multicolumn{1}{c|}{40.12} &
  \multicolumn{1}{c|}{81.12} &
  \multicolumn{1}{c|}{90.20} &
  97.55 \\ \hline
\multicolumn{1}{|c|}{Mobilenet\_v3\_large} &
  \multicolumn{1}{c|}{18.01} &
  \multicolumn{1}{c|}{32.96} &
  \multicolumn{1}{c|}{50.81} &
  \multicolumn{1}{c|}{80.52} &
  \multicolumn{1}{c|}{34.18} &
  \multicolumn{1}{c|}{72.04} &
  \multicolumn{1}{c|}{83.46} &
  96.83 \\ \hline
\multicolumn{1}{|c|}{ResNet34} &
  \multicolumn{1}{c|}{18.10} &
  \multicolumn{1}{c|}{33.90} &
  \multicolumn{1}{c|}{51.23} &
  \multicolumn{1}{c|}{79.50} &
  \multicolumn{1}{c|}{36.93} &
  \multicolumn{1}{c|}{79.38} &
  \multicolumn{1}{c|}{88.77} &
  96.63 \\ \hline
\multicolumn{1}{|c|}{ResNet101} &
  \multicolumn{1}{c|}{16.81} &
  \multicolumn{1}{c|}{32.61} &
  \multicolumn{1}{c|}{52.17} &
  \multicolumn{1}{c|}{79.59} &
  \multicolumn{1}{c|}{\mycc31.05} &
  \multicolumn{1}{c|}{\mycc66.32} &
  \multicolumn{1}{c|}{\mycc80.30} &
  \mycc95.51 \\ \hline
\multicolumn{9}{|c|}{Vision Transformers} \\ \hline
\multicolumn{1}{|c|}{Swin\_v2\_b\_} &
  \multicolumn{1}{c|}{15.57} &
  \multicolumn{1}{c|}{27.15} &
  \multicolumn{1}{c|}{44.15} &
  \multicolumn{1}{c|}{80.70} &
  \multicolumn{1}{c|}{40.40} &
  \multicolumn{1}{c|}{78.45} &
  \multicolumn{1}{c|}{88.16} &
  97.24 \\ \hline
\multicolumn{1}{|c|}{Swin\_v2\_s} &
  \multicolumn{1}{c|}{16.56} &
  \multicolumn{1}{c|}{29.97} &
  \multicolumn{1}{c|}{48.07} &
  \multicolumn{1}{c|}{76.65} &
  \multicolumn{1}{c|}{\mycc33.27} &
  \multicolumn{1}{c|}{\mycc68.67} &
  \multicolumn{1}{c|}{\mycc82.55} &
  \mycc94.79 \\ \hline
\multicolumn{1}{|c|}{Swin\_v2\_t} &
  \multicolumn{1}{c|}{\mycc15.11} &
  \multicolumn{1}{c|}{\mycc25.53} &
  \multicolumn{1}{c|}{\mycc42.27} &
  \multicolumn{1}{c|}{\mycc75.49} &
  \multicolumn{1}{c|}{36.83} &
  \multicolumn{1}{c|}{75.81} &
  \multicolumn{1}{c|}{86.83} &
  96.93 \\ \hline
\end{tabular}%
\end{table*}

\begin{figure*}[]
\centering
\includegraphics[scale=0.4]{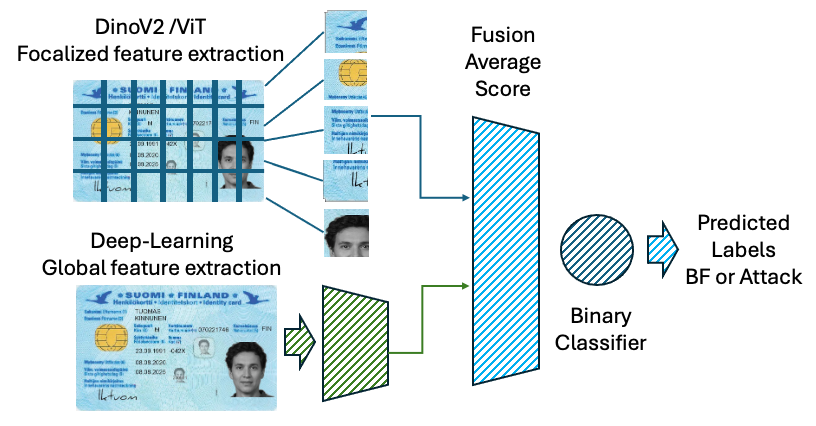}
\caption{ID card Presentation Attack Detection using a fusion pipeline.}
\label{fig:modelo-fusion}
\end{figure*}

\subsection{Experiment 3 - Fine-tuning}

For fine-tuning in the $FM$, the pre-trained models were used to extract the features (embeddings). The sizes of each embedding for the deep learning model and the Vision Transformer were set up according to Experiment 1.  For DinoV2 and CLIP, the size was set to $1\times708$ and $1\times256$, respectively. 

For fine-tuning, we considering added a small head based on a neural network of two-class model in the LOO protocol to identify the most challenging attack. 

All the models were trained using SGD optimisers. Different numbers of layers were explored from $1$ up to $3$. The head with only one layer obtained the best results. The learning rates were analysed using a grid search, ranging from $1e-3$ to $1e-6$, the batch size was set to $64$. The $lr$ value $1e-3$ achieved the best results after $100$ epochs.

Table \ref{tab:finetuning-loo} shows the results for the fine-tuning approach using a LOO protocol, considering one layer in the head as a classifier for the CHL-DB dataset. 

For the foundation model, the best results were obtained by Dinov2-visl14 with an EER of 4.13\% for border, 1.28\% for printed and 2.73\% for screen. Printed and screen again are the most challenging attacks.

Table \ref{tab:LOO-IDnet} shows the results for the fine-tuning approach using a LOO protocol considering one layer in the head as a classifier for the ID-Net dataset. 

The best result was obtained by Dinov2-ViT-l14 with an EER of 11.73\% for Face-replacement, 19.18\% for Face combined, 5.91\% for Face-morphing and 12.36\% for Face copy-move. Face-combined was identified as the most challenging attack for ID-Net. Conversely, a face-morphing attack on an ID card was recognised as the easiest to detect.


\subsection{Benchmarking on open-set database}
Additionally, in order to compare the results and evaluate the generalisation capabilities, we used the ID-Net open-set dataset to assess the performance of all the classifiers with only two classes: bona fide versus border printed and screen attack together. For ID-Net, bona fide versus combined, copy-move, face-morphing and face-replacement attacks together, too.

Table \ref{tabla2c} shows the best results for a two-class classifier in a fine-tuning approach with a small head with one extra layer without the LOO protocol. For all the models, the CHL-DBP dataset obtained the best results in EER in comparison to the ID-Net. This performance shows that the bona fide images are very relevant to separate classes and obtain generalisation capabilities. 

The best results overall were reached by DinoV2-vitl14 with an EER of 6.33\% for CHL-DBP in comparison to the best result on ID-Net with an EER of 27.86\% for DinoV2-vitb14.

For the deep learning methods the best results were reached for DenseNet121 with 15.61\%  CHL-DBP for instead of 31.04\% for ResNet101 with ID-Net.

For the vision transformer, SwinT-v2-t obtained the best result with 15.11\% for CHL-DBP instead of 33.27\% fusing SwinT-v2-s on ID-Net.



\begin{figure*}[]
\centering
\includegraphics[scale=0.26]{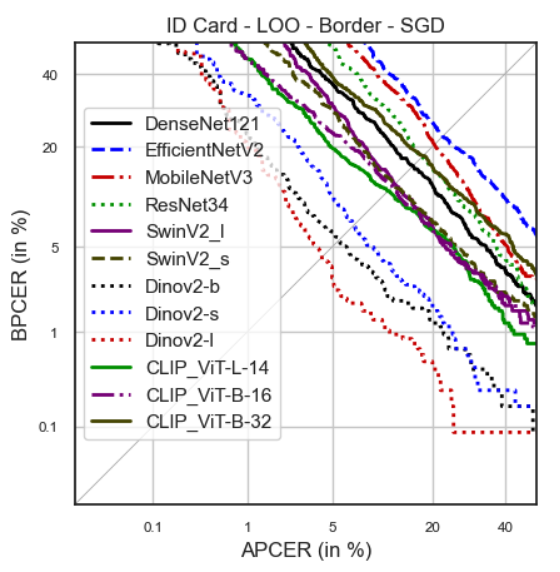}%
\includegraphics[scale=0.21]{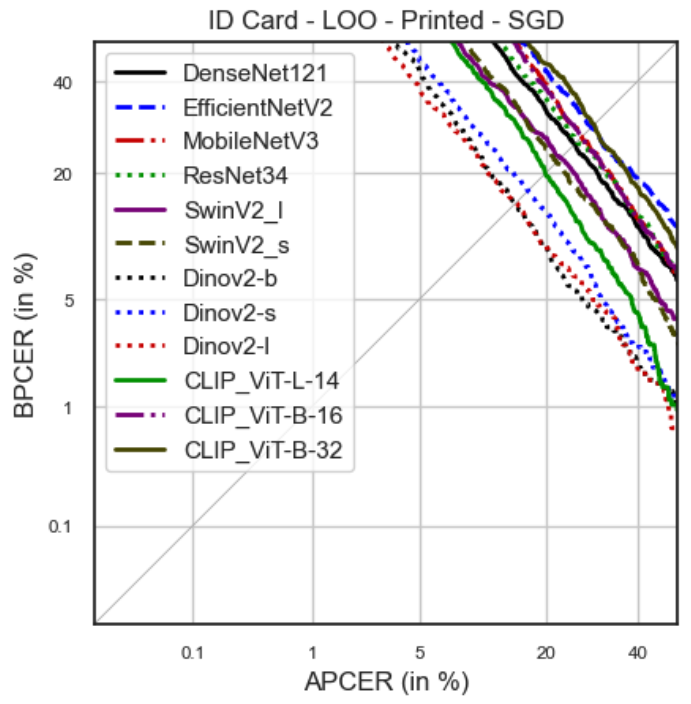}%
\includegraphics[scale=0.27]{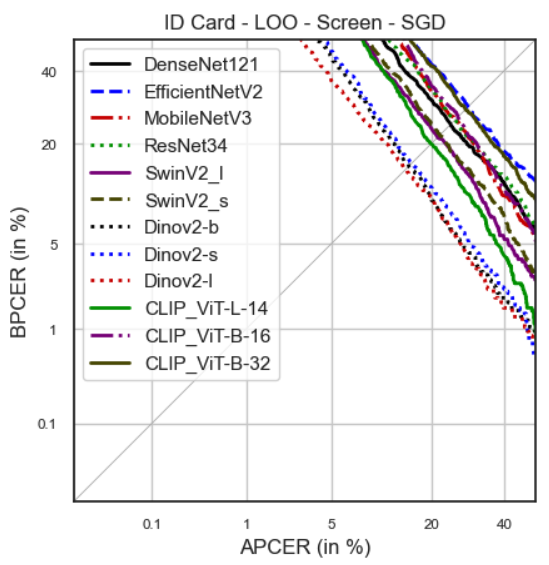}\\
\includegraphics[scale=0.26]{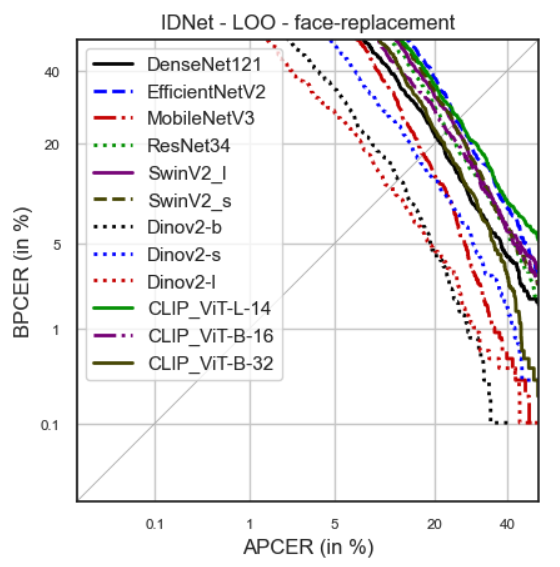}%
\includegraphics[scale=0.27]{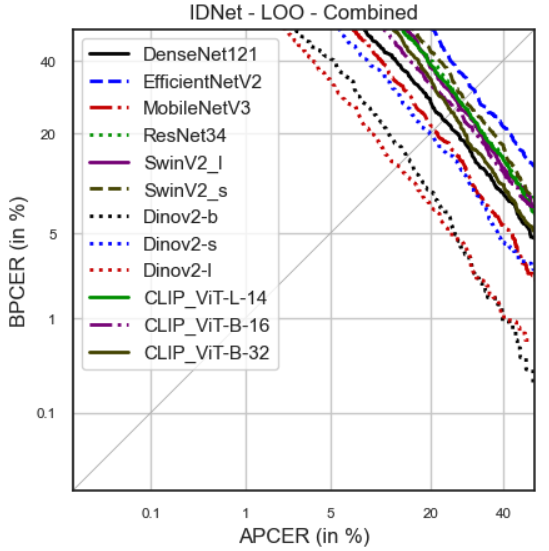}%
\includegraphics[scale=0.27]{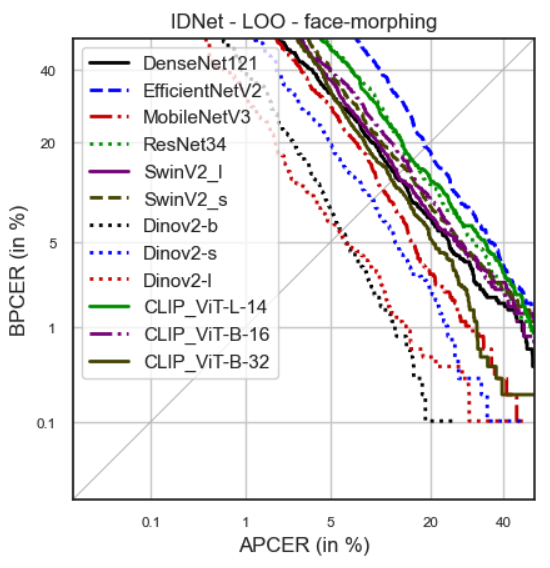}%
\caption{Summary of the DET curves for fine-tuning and Leave-out-protocol for the two datasets considering Deep learning, Vision-Transformer and Foundation models. TOP: CHL-DBP Chilean ID Card(a) Border, Printed and Attack. Bottom: Open-access ID-Net cards, for face-replacement, combined and face-morphing.}
\label{fig_sim}
\end{figure*}

Figure \ref{fig_sim} shows the DET curves summarising the results for the CHL-DBP and open-set datasets (ID-Net) in a LOO protocol. All the network results are deployed to compare deep learning, vision transformer and Foundation models. In all the scenarios, zero-shot and fine-tuning DinoV2 outperform CLIP and the other classifiers.

\subsection{Fusion level}
\label{improve}
Based on the analysis of the proposal's results, we believe it is essential to leverage the benchmarks and outcomes obtained. 
Embedding generation through Vision Transformers (ViT) differs significantly from traditional deep learning models, particularly convolutional neural networks (CNNs). While CNNs process visual data by hierarchically extracting features through convolutional layers, ViTs employ a fundamentally different approach, emphasising attention mechanisms. Vision Transformers operate by dividing images into patches and treating these patches as tokens, similar to how words are treated in natural language processing. This allows them to capture long-range dependencies and contextual relationships across the entire image more effectively than CNNs, which primarily focus on local features. 

\begin{figure}[H]
\centering
\includegraphics[scale=0.22]{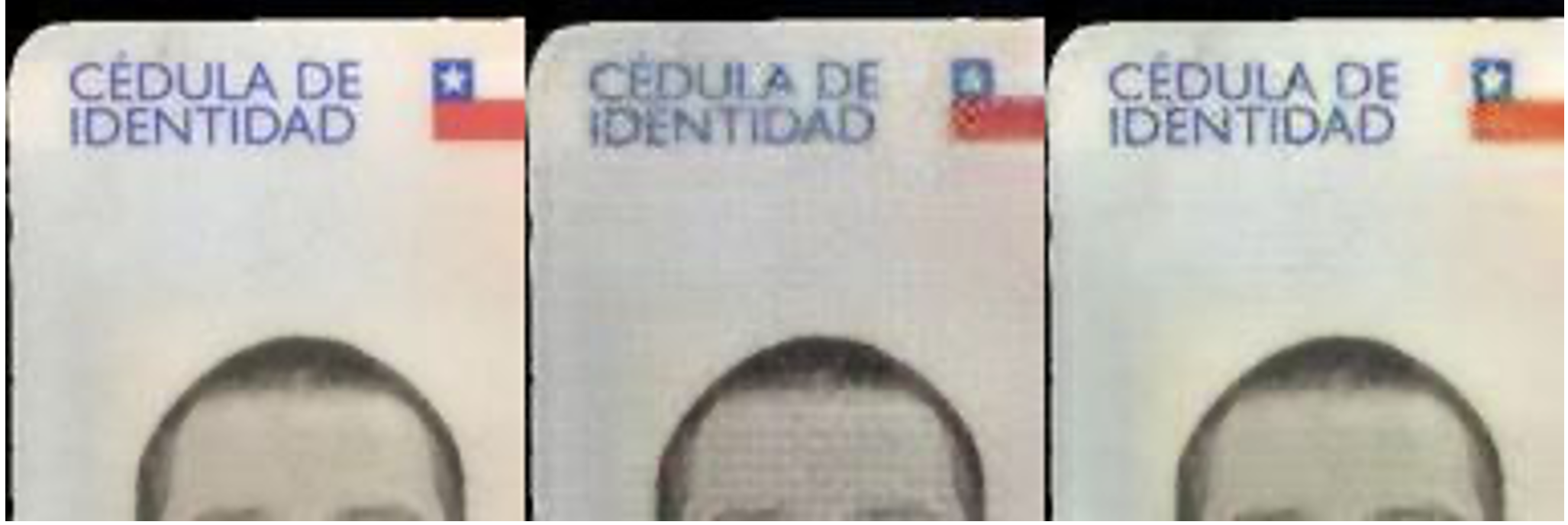}%
\caption{From left to right: images patch sample zoomed for bona fide, printed and screen attacks. Zoom in to see traces/artefact details detected by Dinov2.}
\label{fig:fine-grane}
\end{figure}

Additionally, ViTs rely on self-attention mechanisms to weigh the importance of different patches, enabling them to focus on relevant regions of an image adaptively as is illustrated in Figure \ref{fig:fine-grane}. This enhances their ability to generate rich embeddings that encapsulate spatial and semantic information, making ViTs particularly powerful for various vision-related tasks and applications where the context is crucial.

The ID-Net dataset shows a higher value of EER based on the similarity of the bona fide images with the attack. To enhance the ID-Net results, we suggest integrating the method that achieved the lowest EER from the best-trained model developed from scratch based on deep learning and Imagenet weights, which detects more global features, with the top-performing foundation model based on DinoV2, which focuses on local features, as illustrated in Figure \ref{fig:modelo-fusion}. The features extracted at different levels and variability for the image can boost the results. 
In this approach, we will fuse the features at the score level for both systems and estimate the average between both scores. 

Figure \ref{tab:my-fusion} illustrates the improvements with the approach at the fusion score level. The results previously obtained were reduced from 15.65\% EER obtained for ID-Net to a new EER of 8.25\%. Table \ref{tab:my-fusion} summarises the comparison between the best previous approaches and the new fusion approach.

\begin{table}[H]
\centering
\scriptsize
\caption{Results for fusion approach.}
\label{tab:my-fusion}
\begin{tabular}{|c|ccc|}
\hline
\multirow{2}{*}{\begin{tabular}[c]{@{}c@{}}Model\\ From Table V\end{tabular}} & \multicolumn{3}{c|}{ID-Net Open-set}                            \\ \cline{2-4} 
 &
  \multicolumn{1}{c|}{\begin{tabular}[c]{@{}c@{}}EER\\ (\%)\end{tabular}} &
  \multicolumn{1}{c|}{\begin{tabular}[c]{@{}c@{}}BPCER10\\ (\%)\end{tabular}} &
  \begin{tabular}[c]{@{}c@{}}BPCER20\\ (\%)\end{tabular} \\ \hline
DinoV2-vitb14                                                                 & \multicolumn{1}{c|}{27.86} & \multicolumn{1}{c|}{61.40} & 76.13 \\ \hline
DensetNet121                                                                  & \multicolumn{1}{c|}{15.65} & \multicolumn{1}{c|}{26.21} & 42.52 \\ \hline
\textbf{Fusion}                                                                        & \multicolumn{1}{c|}{\mycc8.25}  & \multicolumn{1}{c|}{\mycc18.25} & \mycc23.21 \\ \hline
\end{tabular}%
\end{table}

\section{Conclusion}
\label{sec:conclusion}

Our experiments show that using a foundation model effectively may help to improve the generalisation capability task based on the limitation of the data available. The results outperform deep learning and traditional vision transformers such as Swin-Transformer. The quantity of data used in each pre-trained method makes the difference, even if foundation models do not have an ID Card in the training set. DinoV2 outperformed CLIP in all the evaluations.

It is essential to highlight that bona fide images make a great difference in performance if we want to improve the results and generalisation capabilities in the different PAD models. The private dataset ID-NET uses as a genuine ID card image an image created from an ID template modified by stable diffusion models; thus, the genuine image is a synthetic image. This data type confuses the classifiers because the attacks are also generated using similar techniques. The model observes no handcrafted attacks, which could be a weakness to basic or handcrafted attacks.

Also, most open-set datasets available in the state of the art are most beneficial for testing the PAD system instead of training. This dataset's creation did not follow any ICAO rules to generate realistic images in the face photo and the text area.

Our proposed approach, based on the fusion of the best from the Foundation model and Deep learning methods, allows us to capture global and local features and detect traces and artefacts on fake ID cards, showing an efficient way to adapt the generalisation capabilities of the foundation models. This saves computational power and allows us to extract and complement the features based on two different models.

As feature work, we must focus on improving techniques oriented to generating "bona fide" images and mimicking their features instead of only focusing on attacks on ID card images. Exploring the text prompt from CLIP may also help to improve the results, together with the inclusion of new models such as BEiT-3, ConvNeXt-V2, and Qwen2.5.

\section*{Acknowledgements}
This research work has been partially funded by the European Union (EU) under G.A. Nº 101121280 (EINSTEIN) and CarMen (101168325), and the German Federal Ministry of Education and Research and the Hessian Ministry of Higher Education, Research, Science and the Arts within their joint support of the National Research Center for Applied Cybersecurity ATHENE.
\newpage

\newpage
\bibliographystyle{IEEEtran}
\bibliography{sample.bib}

\newpage
\begin{IEEEbiography}[{\includegraphics[width=0.9in,height=1.25in,clip,keepaspectratio]{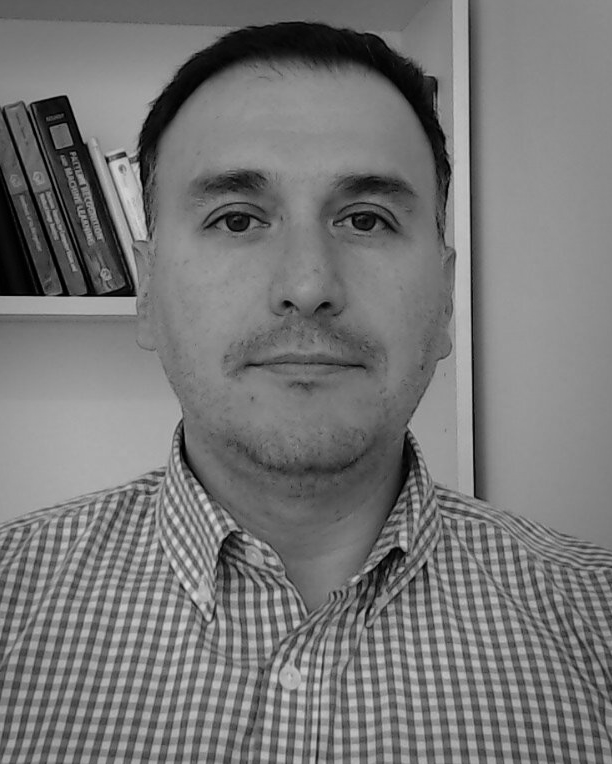}}]{Juan Tapia} received the P.E. degree in electronics engineering from Universidad Mayor, in 2004, and the M.S. and Ph.D. degrees in electrical engineering from the Department of Electrical Engineering, Universidad de Chile, in 2012 and 2016, respectively. In addition, he spent one year of internship with the University of Notre Dame. In 2016, he received the Award for Best Ph.D. Thesis. From 2016 to 2017, he was an Assistant Professor at Universidad Andres Bello. From 2018 to 2020, he was the Research and Development Director for the electricity and electronics area with INACAP, Universidad Tecnologica de Chile, the Research and Development Director of TOC Biometrics Company, and an International Advisor on biometrics for face, iris applications and forensic/tampering ID-card detection. He is currently an Entrepreneur and a Senior Researcher with Hochschule Darmstadt (HDA), leading EU projects, such as iMARS, EINSTEIN and CarMen. His main research interests include pattern recognition and deep learning applied to iris biometrics, morphing, feature fusion, and feature selection. He serves as a reviewer for a number of journals and conferences. He is on behalf of the German DIN Member of the ISO/IEC Sub-Committee 37 on biometrics.
\end{IEEEbiography}
\vspace{-0.3cm}

\begin{IEEEbiography}[{\includegraphics[width=0.9in,height=1.25in,clip,keepaspectratio]{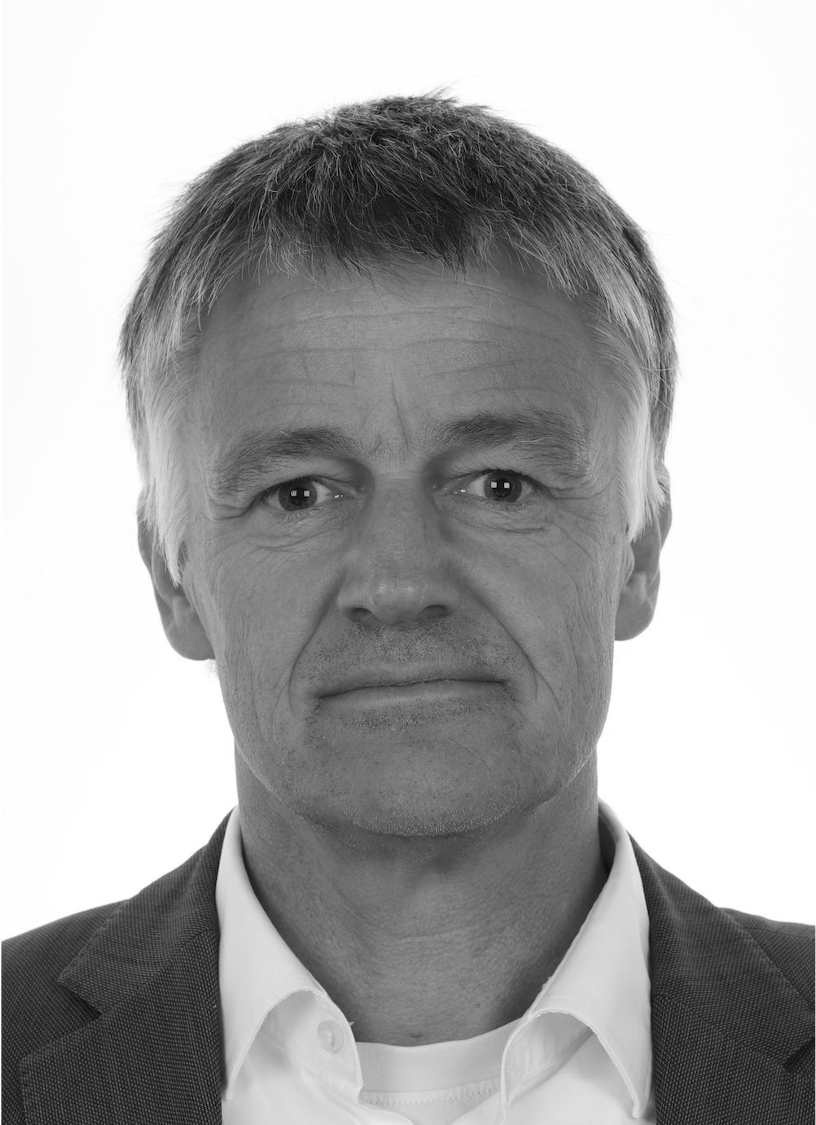}}]{Christoph Busch} is a member of the Department of Information Security and Communication Technology (IIK) at the Norwegian University of Science and Technology (NTNU), Norway. He holds a joint appointment with the computer science faculty at Hochschule Darmstadt (HDA), Germany. Further, he lectures the course Biometric Systems at Denmark’s DTU since 2007. On behalf of the German BSI he has been the coordinator for the project series BioIS, BioFace, BioFinger, BioKeyS Pilot-DB, KBEinweg and NFIQ2.0. In the European research program, he was the initiator of the Integrated Project 3D-Face, FIDELITY and iMARS. Further, he was/is partner in the projects TURBINE, BEST Network, ORIGINS, INGRESS, PIDaaS, SOTAMD, RESPECT and TReSPAsS. He is also principal investigator at the German National Research Center for Applied Cybersecurity (ATHENE). Moreover Christoph Busch is co-founder and member of the board of the European Association for Biometrics (www.eab.org) which was established in 2011 and assembles in the meantime more than 200 institutional members. Christoph co-authored more than 700 technical papers and has been a speaker at international conferences. He is a member of the editorial board of the IET journal.
\end{IEEEbiography}

\vfill

\end{document}